\documentclass[journal]{IEEEtran}
\usepackage{rotating}
\usepackage{booktabs}
\usepackage{hyperref}
\usepackage{url}
\usepackage[hyphenbreaks]{breakurl}

\ifCLASSINFOpdf
  \usepackage{graphicx}
\else

\fi

\usepackage{amsmath}

\usepackage{array}

\begin{document}

\title{Reliability-based cleaning of noisy training labels with inductive conformal prediction in multi-modal biomedical data mining}

\author{{Xianghao Zhan, Qinmei Xu, Yuanning Zheng, Guangming Lu, Olivier Gevaert}
\thanks{Xianghao Zhan is with the Department of Bioengineering, Stanford University, Stanford, 94305, USA. (e-mail: xzhan96@stanford.edu).}
\thanks{Qinmei Xu, Yuanning Zheng, and Olivier Gevaert are with the Department of Biomedical Data Science (BMIR), Department of Medicine, Stanford University, Stanford, CA 94304, USA. (e-mail: xqm0629@stanford.edu; eric2021@stanford.edu; ogevaert@stanford.edu).}
\thanks{Guangming Lu is with the Department of Medical Imaging, Jinling Hospital, Nanjing, Jiangsu, China. (e-mail: cjr.luguangming@vip.163.com).}
\thanks{Corresponding authors: Olivier Gevaert: ogevaert@stanford.edu.}}%

\markboth{}%
{Shell \MakeLowercase{\textit{et al.}}: Bare Demo of IEEEtran.cls for IEEE Journals}

\maketitle

\begin{abstract}
Accurately labeling biomedical data presents a challenge. Traditional semi-supervised learning methods often under-utilize available unlabeled data. To address this, we propose a novel reliability-based training data cleaning method employing inductive conformal prediction (ICP). This method capitalizes on a small set of accurately labeled training data and leverages ICP-calculated reliability metrics to rectify mislabeled data and outliers within vast quantities of noisy training data. The efficacy of the method is validated across three classification tasks within distinct modalities: filtering drug-induced-liver-injury (DILI) literature with title and abstract, predicting ICU admission of COVID-19 patients through CT radiomics and electronic health records, and subtyping breast cancer using RNA-sequencing data. Varying levels of noise to the training labels were introduced through label permutation. Results show significant enhancements in classification performance: accuracy enhancement in 86 out of 96 DILI experiments (up to 11.4\%), AUROC and AUPRC enhancements in all 48 COVID-19 experiments (up to 23.8\% and 69.8\%), and accuracy and macro-average F1 score improvements in 47 out of 48 RNA-sequencing experiments (up to 74.6\% and 89.0\%). Our method offers the potential to substantially boost classification performance in multi-modal biomedical machine learning tasks. Importantly, it accomplishes this without necessitating an excessive volume of meticulously curated training data.

\end{abstract}


\IEEEpeerreviewmaketitle

\section{Introduction}
Machine learning, particularly supervised learning, has contributed to many successful applications in the biomedical field, such as clinical decision support systems to predict sepsis based on electronic health records \cite{desautels2016prediction}, to triage COVID-19 patients based on predicted needs of intensive care units and mechanical ventilation based on computed tomography (CT) radiomics and electronic health records \cite{xuzhan2021ai}, to predict patient prognosis from multi-omics data \cite{steyaert2023multimodal,zheng2023spatial}, and information retrieval system based on natural language processing (NLP) to rapidly extract structured clinical diagnosis codes and symptoms from free-text clinical notes \cite{53_NLP_clinical_texts}. 

However, the challenge in obtaining accurately labeled data with high-fidelity in the multi-modal biomedical machine learning has posed a significant challenge in developing effective models and algorithms. For instance, to create accurately labeled diagnoses and lesion segmentation, experienced radiologists have to manually process a large number of radiological images (e.g. X-ray, CT) \cite{xuzhan2021ai,xu2021advanced,HuangXu2023distinguishing}. This process typically involves a group of independent radiologists to label randomized and blinded images, which can pose a huge burden for both clinicians if they are required to provide labels with high confidence and fidelity. Similarly, to train an effective model for biomedical natural language processing, experts need to read through thousands of free-text notes to generate text classifications \cite{zhan2022filter}. This text labeling process can take a long time but without high quality labels, it can be difficult to train models. As a case in point, labeling noise of diagnostic codes have become a recognized issue in cardiovascular patients' electronic health records \cite{53_NLP_clinical_texts,mccarthy2019,chang2016,goldstein1998accuracy}.

To address the scarcity of well-curated labels in biomedical data, weakly supervised learning and semi-supervised learning have been proposed. Weakly supervised learning leverages noisy labels generated by weak labeling based on rules and benefits from the information from multiple noisy labels, which has been shown to be effective in multiple tasks \cite{hassan2022supervised,yang2020weakly,jain2016weakly,prest2011weakly}. Semi-supervised learning generally uses both labeled data and unlabeled data \cite{qi2020small,adeli2018semi,ines2021biomedical,ge2020deep,53_label_propagation}, leverages the distributional information in the labeled data and uses the unlabeled data to boost the model performance \cite{qi2020small,yamada2022guiding}. Both mechanisms have been shown to be successful in multiple applications but there are still limitations. Weakly supervised learning mechanism may not fully benefit from the ground-truth labels made by clinicians or radiologists with high confidence. It also requires an additional step of weak labeling and generation of noisy labels, which can be dependent on the expert knowledge that may not be generalizable across different research fields \cite{zhou2018brief}. It might also be challenging to determine what the incomplete or weakly labeled data represents and this ambiguity can lead to inaccurate learning \cite{jain2009active}. When the labeling process is weak, it may be difficult to measure the performance of a model accurately as well. The model evaluation metrics might be unreliable \cite{chapelle2009semi}. As an example, for diagnostic code extraction from clinical notes, the weak labeling process in weakly supervised learning still takes time and needs some expert knowledge to design rule-based information extraction systems based on regularized expression. Traditional semi-supervised learning, on the other hand, may lose some of the information of the data which are deemed as the unlabeled data in the unsupervised learning part. For example, radiologists may have knowledge of how to label a portion of unlabeled X-ray images but the confidence might be low. In this case, if the data are deemed as labeled data for the supervised learning part, they contain noise in the labels. On the contrary, if they are deemed as unlabeled data, the radiologists' understanding and prior knowledge are not fully made use of. However, even though there could be labeling noise in the labels given by the radiologists (i.e. wrongly labeled data), the original labels may still contain information. Therefore, regarding them as definitely correct reference labels or absolutely unlabeled data may not be the best-optimized idea. Additionally, semi-supervised learning often makes the assumption that the unlabeled data follows a similar distribution as the labeled data (e.g., the labeled data and unlabeled data cover similar space on a manifold or Euclidean feature space). If this assumption is incorrect, then the model's performance may be degraded \cite{chapelle2009semi}. 


The complexity of weakly supervised learning and the under-utilization of noisy labels and strict distribution assumptions in traditional semi-supervised learning lead us to think whether we can further benefit from a small portion of well-curated high quality data while also using a large portion of labeled data with noise. If we can accurately label certain portions of training data, can we use this information to clean the other part of potentially noisy data? If we can correct noisy training data, can we further improve the performance in supervised learning tasks?

One method to curate the noisily labeled data is to use the reliability quantified by conformal predictions \cite{vovk_algorithmic_2005,angelopoulos2021gentle}. Based on a weak assumption of independently and identically distributed (i.i.d) data, conformal prediction quantifies the reliability of a particular combination of feature and label based on the calibration of the nonconformity measure \cite{vovk_algorithmic_2005,angelopoulos2021gentle,zhan2020electronic}. In previous studies, conformal prediction has been shown effective in reliability-based unlabeled data augmentation tasks \cite{liu2021boost,liu2022cpsc,zhan2018online}. In these studies, researchers start with a labeled training data set and train conformal predictors to filter the unlabeled data. Researchers attach pseudo-labels to the unlabeled data based on certain models, and remove the unreliable pseudo-labeled data. The augmented training data set can lead to statistically significantly better classification performance with robustness. The reliability-based unlabeled data augmentation method also showed effectiveness in improving model generalizability under domain drifts between the training and validation/test datasets \cite{liu2021boost,wang2022unsupervised}. The "training data augmentation" idea in these works inspired us to think in the opposite direction with the idea of "trimming training data": can we use the inductive conformal prediction (ICP) framework to clean the noisy training data, to improve the training data quality and thus to improve the classification performances in the downstream tasks? The setup of the ICP framework corresponds naturally to the scenarios we face: it requires a small portion of calibration data and a large portion of proper training data \cite{vovk_algorithmic_2005,angelopoulos2021gentle,liu2021boost}. The latter is used to train a conformal predictor to compute the nonconformity measure while the former is used to calibrate the reliability of a particular combination of a feature and a label based on the reference distribution of nonconformity measure on the calibration set. To leverage the ICP, we can deem the well-curated labeled training data as the calibration set while using the large portion of remaining data with unknown label quality as the proper training set. Instead of the typical usage of ICP to quantify the test samples' reliability, we use the ICP to quantify the proper training samples' reliability in a retrospective manner and detect whether the original labels are of high reliability based on the P-values, or more specifically, how well the original label conforms to the reference distribution of nonconformity measure on the calibration set and whether there is another label showing better conformity.

In this study, we propose a reliability-based training data cleaning method based on ICP to detect potentially wrongly labeled data (another label is more likely to be correct) and outliers (the out-of-distribution data, no labels are likely to be correct) and make corrections based on rules (Methods Section 2, 3). We validate the effectiveness of the method on three biomedical machine learning classification tasks with different modalities: a clinical NLP task, a radiomics task using quantitative image features extracted from radiographic images, an electronic health record task, and an RNA-sequencing task (see Methods Section 1). We manually pollute the training data to simulate different levels of labeling noises in the training data, by manually permuting different percentages of labels. The classification performance metrics are evaluated with and without training data cleaning (see Methods Section 4). Additionally, the mechanisms of the cleaning method were investigated by visualizing the cleaning processes: how many corrections of wrongly labeled data and how many removals of outliers were made by the method under different hyper-parameters.

\section{Methods}
\subsection{Datasets and task description}
To investigate a broad range of medical data modalities and different scales of feature dimensionality, in this study, we tested the reliability-based training data cleaning method based on inductive conformal prediction on three classification tasks: 1) a natural language processing task: filter drug-induced liver injury (DILI) literature based on word2vec (W2V) and sent2vec (S2V) embeddings \cite{zhan2022filter}; 2) an imaging and electronic health record task: predict whether a COVID-19 patient in the general ward will be admitted to the intensive care unit (ICU) \cite{xuzhan2021ai}; 3) an RNA-sequencing (RNA-seq) task: classify breast cancer subtypes based on The Cancer Genome Atlas Program (TCGA) RNA-seq dataset \cite{cancer2012comprehensive}. The details of these datasets are introduced in Table \ref{datasets} and the following paragraphs.

\begin{table*}
    \small
    \centering
    \caption{The basic information of datasets used in this study.}
    \label{tab:1}       
    \begin{tabular}{cccccccc}
        \hline\noalign{\smallskip}
        Dataset & Modality & Task type & Features & Dimensionality & \# of samples & Metrics & Ref. \\
        \noalign{\smallskip}\hline\noalign{\smallskip}
        DILI & Text & Binary & W2V, S2V & 200, 700 & 14,203 & Accuracy & \cite{zhan2022filter}\\
        & & & &  & (+: 7,177, -: 7,026) & \\
        COVID-19 & Imaging, EHR & Binary & Radiomics, & 9,940 & 2,113 & AUROC, AUPRC & \cite{xuzhan2021ai} \\
            & & & lab and clinical data &  & (+: 60, -: 2,053) & \\
        TCGA & RNA-seq & Multi-class & RNA-seq & 31,098 & 1,096 & Accuracy, F1 score & \cite{cancer2012comprehensive} \\	
        \hline\noalign{\smallskip}
    \end{tabular}
    \label{datasets}
\end{table*}

The DILI dataset was released by the Annual International Conference on Critical Assessment of Massive Data Analysis (CAMDA 2021), with the DILI-positive samples (7,177) and DILI-negative samples (7,026) curated by FDA experts. The data involve both title and abstract of publications and the task is to predict whether a publication contains DILI information. The detailed pre-processing of the text can be found in the previous publication \cite{zhan2022filter}: the text were lowercased, with additional removals of punctuation, numeric, special characters, multiple white spaces, stop words, and the text was finally tokenized with Gensim library on Python 3.7 \cite{rehurek_software_2010}. Then, we generated the text embeddings based on the biomedical sent2vec (S2V) model \cite{pagliardini_unsupervised_2018} and biomedical word2vec (W2V) model \cite{zhang_biowordvec_2019} because they have shown good classification performances on this challenge in previous studies with logistic regression classifiers, and these two text vectorizations generate text embeddings with low feature dimensionality (700 for S2V and 200 for W2V) \cite{zhan2022filter}.

COVID-19 data (n=2,113) in this study were collected from 40 hospitals in China from December 27, 2019 to March 31, 2020 \cite{xuzhan2021ai}. Patients selection followed the inclusion criteria: (a) RT-PCR confirmed positive severe acute respiratory syndrome coronavirus (SARS-CoV-2) nucleic acid test; (b) baseline chest CT examinations and laboratory tests on admission; (c) short-term prognosis information (discharge or admission to ICU). Data for each patient included: 1) Clinical data based on electronic Health Records (EHR): (a) demographics: age and gender; (b) comorbidities: coronary heart disease, diabetes, hypertension, chronic obstructive lung disease (COPD), chronic liver disease, chronic kidney disease, and carcinoma; (c) clinical symptoms: fever, cough, myalgia, fatigue, headache, nausea or vomiting, diarrhea, abdominal pain, and dyspnea on admission. 2) Lab data based on laboratory test: (a) blood routine: white blood cell (WBC) count ($\times 10^9/L$), neutrophil count ($ \times 10^9/L$), lymphocyte count ($\times 10^9/L$), platelet count ($\times 10^9/L$), and hemoglobin ($g/L$); (b) coagulation function: prothrombin time (PT) ($s$), activated partial thromboplastin time (aPTT) ($s$), and D-dimer ($mg/L$); (c) blood biochemistry: albumin ($g/L$), alanine aminotransferase (ALT) ($U/L$), aspartate Aminotransferase (AST) ($U/L$), total bilirubin ($mmol/L$), serum potassium ($mmol/L$), sodium ($mmol/L$), creatinine ($\mu mol/L$), creatine kinase (CK) ($U/L$), lactate dehydrogenase (LDH) ($U/L$), $\alpha$-Hydroxybutyrate dehydrogenase (HBDH) ($U/L$); (d) infection-related biomarkers: C-reactive protein (CRP) ($mg/L$). 3) Radiomics data based on CT imaging: a commercial deep-learning AI system (Beijing Deepwise \& League of PhD Technology Co. Ltd) was first used to detect and segment the pneumonia lesion, and two radiologists checked the results of the automatic segmentation. Then, pyradiomics (v3.0) running in the Linux platform was adopted to extract radiomic features (1,652 features per lesion). Next, for a given patient and for each radiomic feature, we summarized the distribution of the feature values across all the lesions for the patient by several summary statistics (mean, median, standard deviation, skewness, the first quartile, the third quartile) and the number of lesions. Finally, a total of 9,913 quantitative radiomic features were extracted from CT images for each patient. Detailed data collection and preprocessing are shown in the supplementary materials.


RNA-seq data of the TCGA breast cancer cohort \cite{cancer2012comprehensive} were obtained from the UCSC Xena browser \cite{goldman2020visualizing}. Gene expression values were normalized using the Fragments Per Kilobase of transcript per Million mapped reads (FPKM) method. The dataset included 1,096 patients, and the molecular subtypes were labeled using the methods as described previously \cite{thennavan2021molecular}. The number in each subtype was: LumA (n = 569), LumB (n = 219), Basal (n = 187), HER2 (n = 82), and normal-like (n = 39). The feature dimensionality is 31,098.

\subsection{Inductive conformal prediction}
\subsubsection{General Pipeline of Inductive Conformal Prediction}
Inductive Conformal Prediction (ICP) is a computational framework that operates under the assumption of identically and independently distributed data. For comprehensive discussions on conformal predictors, readers are directed to past works \cite{vovk_algorithmic_2005,angelopoulos2021gentle,zhan2018online}. To succinctly describe the ICP framework, it begins with splitting the training data into a proper training set and a calibration subset. Nonconformity measures are then derived from the proper training set using specific heuristic rules or algorithms, such as using the conditional probability from a classification model like Conformal Prediction with Shrunken Centroids (CPSC) \cite{liu2022cpsc} or using the ratio of cumulative distances between dissimilar and similar samples as in Conformal Prediction with k-nearest neighbors (CPKNN) \cite{zhan2020electronic,zhan2018online}. Following this, nonconformity measures for the calibration and test data are computed and utilized as calibration statistics. For the calibration set, these nonconformity measures are computed using both features and ground-truth labels. In contrast, for the test data, the nonconformity measures are computed for every possible label within the label space in order to quantify and compare the conformity of each possible label. ICP then uses the empirical distribution of the nonconformity measures on the calibration set as a reference, assessing at which percentile the test sample-label combination's nonconformity measure falls. Based on this percentile, ICP calculates the P-value, which indicates the degree of conformity of a specific feature-label combination to the underlying data distribution, i.e., how well the test feature-label combination conforms to the distribution of the nonconformity measure on the calibration set. This enables the quantification of prediction reliability by considering the conformity of the most likely label to the training data distribution.
	
The ICP frame work used in this study is visualized in top-left section of Fig. \ref{pipeline}, where the entire training set with $b$ samples is divided into a proper training set $\left\{X_{1}, X_{2}, X_{3}, \ldots, X_{a}\right\}$ and a calibration set $\left\{X_{a+1}, X_{a+2}, X_{a+3}, \ldots, X_{b}\right\}(a<b)$. Based on a nonconformity measure algorithm trained on the proper training set, the nonconformity measure of calibration set and proper training set ($A\_Cal$ and $A\_Pro$) are calculated. The P-values, which are the calibrated results reflecting the conformity of a feature-label combination, are then calculated for all possible labels of each sample in the proper training set, and the P-values indicate the reliability of a proper training sample when attached every possible label judged by the nonconformity measure distribution on the calibration set. In this study, to clean the training data with noisy labels, the proper training data represent the majority part of the labeled data but may be noisy with outliers and wrongly labeled data unknown to the users, while the calibration set are the smaller portion of labeled data with well-curated labels. It should be mentioned that different from the typical use case of ICP where it uses the calibration set to compute the P-values for the test set, in this study, we aim to compute the P-values for the samples in proper training set and the leverage the P-values for training data cleaning.

The nonconformity measure for the $i$-th sample with a supposed label $y$, expressed as $\alpha_{i}^{y}$, is determined via the CPSC algorithms, which we will cover in the following subsection. The computation of P-values for a given sample $x_i$ proceeds as outlined below. In this context, $p_{i}^{y}$ signifies the P-value of $x_i$ for a possible label $y$ within the label space. Similarly, $\alpha_{j}^{y}$ represents the nonconformity measure for the $j$-th sample, associated with the label $y$ in the calibration set. Lastly, $\alpha_{i}^{y}$ stands for the nonconformity measure for a potential label $y$ tied to the $i$-th sample in the proper training set. Here, Laplace smoothing was applied.
\begin{equation}
    p_{i}^{y}=\frac{| \{j=a+1, \ldots, b \}:  \alpha_{j}^{y} \geq \alpha_{i}^{y} \mid+1}{b-a+1}
\end{equation}

\subsubsection{Nonconformity Measure Algorithm with Shrunken Centroids}
In this study, we applied our previously developed CPSC algorithms with the shrunken centroids (SC) as the basis for nonconformity measure because it has shown higher computation efficiency and effectiveness in reliability quantification when compared with the conventional conformal prediction with k-nearest neighbors algorithm (CPKNN) and several other conformal predictors \cite{liu2022cpsc}, and its effectiveness in quantifying reliability has been validated on both the DILI dataset \cite{zhan2022filter} and the COVID-19 patient ICU admission prediction task \cite{xu2022tri}.
The computation is based on the following steps similar to the steps of shrunken centroids algorithm \cite{shrunken_centroid} (assume the dimensionality of feature space is denoted as $D$):

In the feature space, we firstly calculate the class centroids $\bar{x}_{m} \in \rm I\!R^{D}$ for class $1,2,...,M$) and the overall centroid $\mu$. ${C}_{m}$ refers to the set of samples in class $m$, ${n}_{m}$ denotes the total number of samples within class $m$ and $x_{jm}$ denotes the $j$-th sample of class $m$. 
\begin{equation}
\label{class centroid}
    \bar{x}_{m}=\sum_{x_{jm} \in C_m}^{} \frac{x_{j m}}{n_{m}}
\end{equation}
\begin{equation}
\label{overall centroid}
\mu=\sum_{i=1}^{n} \frac{x_{i}}{n}
\end{equation}

Then, the pooled standard deviation is computed and the contrasts between class centroids and the overall centroid are normalized using the pooled standard deviation:

\begin{equation}
\label{eq:std_update}
\sigma^{2}=\frac{1}{n-M} \sum_{m=1}^{M} \sum_{x_j \in C m}\left(x_{j}-\bar{x}_{m}\right)^{2}
\end{equation}
\begin{equation}
d_{m}=\left(\bar{x}_{m}-\mu\right) / \sigma
\end{equation}

Next, these contrasts are shrunken towards the overall centroid with a soft threshold symbolized by $\Delta$, which is regarded as a hyperparameter in this study (\ref{shrink_bias}):

\begin{equation}
\label{shrink_bias}
\begin{array}{c}
d_{m}^{\prime}=\operatorname{sign}\left(\mathrm{d}_{m}\right)\left(\lvert\mathrm{d}_{m}\rvert-\Delta\right)_{+} \\h_{+} \quad=\left\{\begin{array}{ll}h & h>0 \\0 & h \leq 0
\end{array}\right.
\end{array}
\end{equation}

The impact of regularization is governed by the threshold parameter $\Delta$. If the absolute value of ${d}_{{x}{y}}$, denoting the contrast in an $x$-th feature for class $y$, is less than the threshold $\Delta$, it is concluded that the related feature lacks sufficient discriminatory power for classification. Consequently, the shrunken contrast for this attribute is reduced to zero, thus discarding the attribute considered as non-contributory and diminishing the dimensionality of the data. Following this, it becomes possible to recalibrate the class centroids taking into account the shrunken contrasts (\ref{eq:adjusted_biases}).

\begin{equation}
\label{eq:adjusted_biases}
\bar{x}_{m}^{\prime}=\mu + s*d_{m}^{\prime}
\end{equation}
\indent

Then, in the feature space, the discriminatory score of a new sample $x^{*}$ can be compared with shrunken centroids for each class:
\begin{equation}
    \label{eq:new_cluster}
\delta_{m}\left(x^{*}\right)=\log \pi_{m}-\frac{1}{2} \sum_{k=1}^{\mathrm{D}} \frac{\left(x_{k}^{*}-\bar{x}_{k m}^{\prime}\right)^{2}}{s_{j}^{2}}
\end{equation}

where $x_{k}^{*}$ denotes the $k$-th feature of the new sample $x^{*}$ while $\bar{x}_{k m}^{\prime}$ denotes the $k$-th value of the shrunken centroid of class $m$.

The discriminatory score $\delta_{m}\left(x^{*}\right)$ quantifies the proximity of a novel sample $(x^{*}$ to the $m$-th shrunken centroid), or in other words, it represents the log probability of $x^{*}$ being part of class $m$, without normalization. This resulting score is constituted by two components: the initial term, $\log \pi_{m}$, indicates the prior probability of class $m$, determined by the frequency of samples in class $m$ amidst all observations. The secondary term constitutes the standardized squared distance between the $m$-th centroids and the new sample. Consequently, the log probability of a specified class $m$, devoid of normalization, $P(Y=m|X)$, is influenced by both the initial class distribution and the sample's closeness to varying centroids.

The probability of the new sample $x^{*}$ from class $k$ can then be modeled with the discriminatory score for all classes:
\begin{equation}
\label{eq:softmax_function}
\hat{p}\left(k | x^{*}\right)=\frac{e^{\delta_{k}\left(x^{*}\right)/T}}{\sum_{\ell=1}^{K} e^{\delta_{\ell}\left(x^{*}\right)/T}}
\end{equation}

In order to obtain a normalized probability distribution (one that spans from 0 to 1 and whose elements sum to 1), the softmax function is employed. This transforms log probabilities (not normalized) from arbitrary real numbers into normalized probabilities. It operates on $\delta_{m}\left(x^{*}\right)$ in a manner comparable to its handling of logits in neural networks \cite{softmax_MIT}. Moreover, a scaling factor $T$ (denoted as 'temperature') is implemented to diminish the value of $\delta_{m}(x^{*})$ with the aim of rendering the predicted probability distribution more uniform or "softer", while conserving the relative probability ranks across each class. The concept of temperature hyperparameter $T$ is adopted from the softmax function used in knowledge distillation. This typically results in a more evenly spread distribution across various labels, hence retaining the information of less probable labels while also mitigating overfitting to some degree \cite{knowledge_distillation}. Simultaneously, as per (\ref{eq:new_cluster}), the posterior probability as given by $\delta_{m}\left(x^{*}\right)$ is invariably a negative value. This could lead to the term $e^{\delta_{m}\left(x^{}\right)}$ in the softmax function becoming exceedingly small and potentially causing numerical instability. This risk is particularly pronounced when handling high-dimensional data, as crucial probability information might be lost in these spiky conditional probablities. By scaling the original $\delta_{m}\left(x^{*}\right)$ with $T$, the absolute value of the exponential term in the softmax function increases, leading to greater information retention. To mitigate the risks of overfitting and numerical instability, $T$ will be tuned as a hyperparameter in this study.

Finally, we convert the predicted probability to a nonconformity measure in ICP framework based on (\ref{eq:prob_to_measure}). Here, we applied a design of the nonconformity measure $\alpha_j^{y_i}$ that has been validated in multiple machine learning applications \cite{zhan2022filter,wang2022unsupervised,xu2022tri}:
\begin{equation}
\label{eq:prob_to_measure}
\alpha_j^{y_i}=0.5-\frac{\hat{p}\left(y_i \mid x_j\right)-\max \hat{p}_{y!=y_i} \left(y_{i} \mid x_{j}\right)}{2}
\end{equation}

\subsection{Training data cleaning methods based on inductive conformal prediction}
In this study, the cleaning method is visualized in Fig. \ref{pipeline}. Upon partitioning the entire dataset into training  (60\%), validation  (20\%) and test sets (20\%), we partition our training data into a proper training set with unknown label qualities (80\% entire training samples) and a well-curated calibration set (20\% entire training samples). Here, to simulate the real-world scenarios in biomedical machine learning applications, we let the proper training set represent the large portion of training samples with unknown label qualities, while the calibration set represent the small portion of training data with clean labels. As is introduced in the previous subsection, ICP can be used to quantify the conformity of a particular feature-label combination based on the well-curated calibration set. In this study, instead of quantifying the conformity for samples in the test set, we aim to quantify the conformity of the noisy training data in the proper training data with the well-curated calibration set. Then, the wrongly labeled data and outliers can be detected and corrected based on the following rules (shown in the bottom left section of Fig. \ref{pipeline}):

\begin{figure*}[!htp]
\centering
\includegraphics[width=0.8\linewidth]{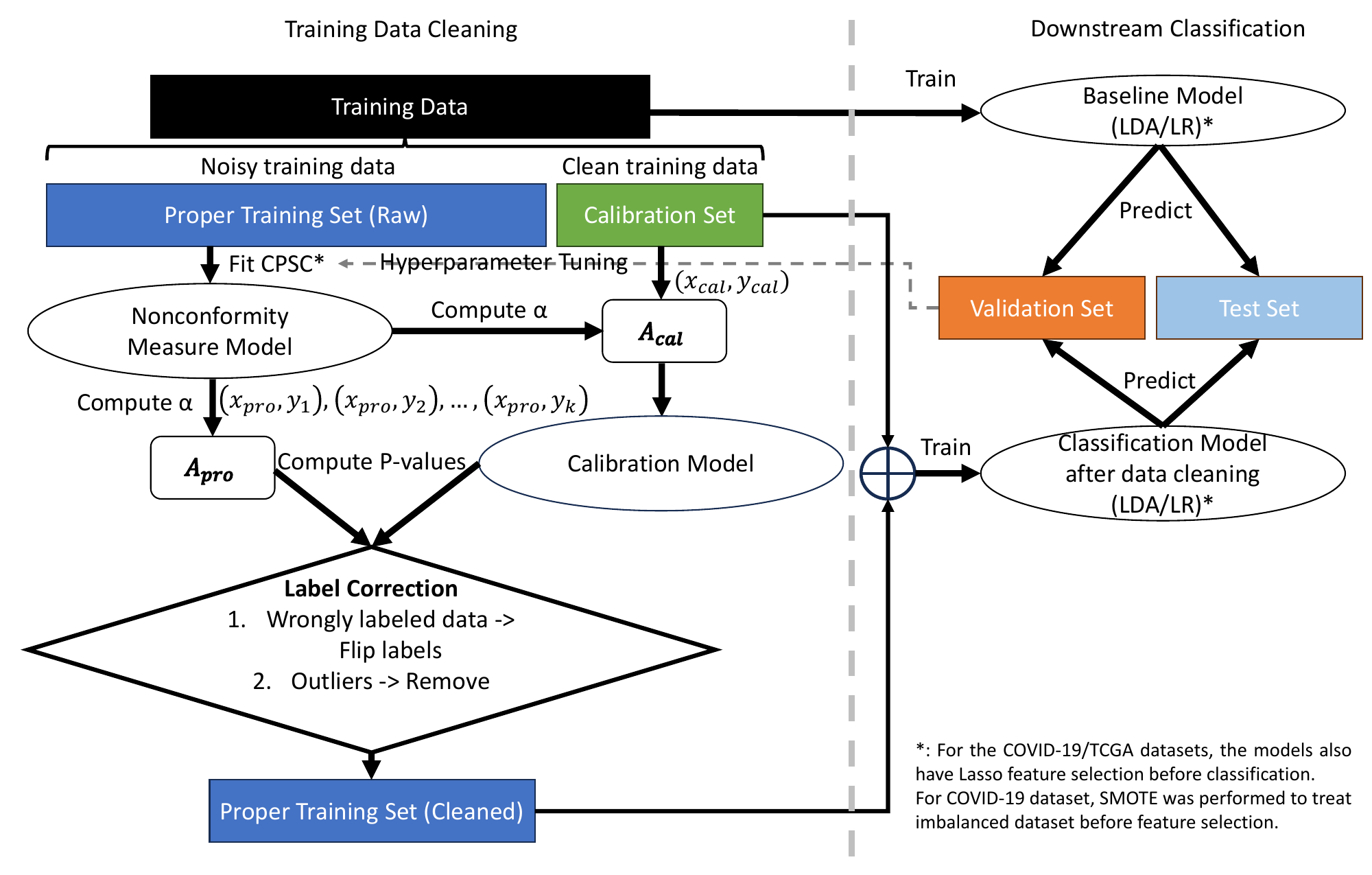}
\caption{The pipeline of the design of the reliability-based training data cleaning method based on inductive conformal prediction and the validation process. The training data cleaning method based on the conformal prediction is shown on the left half of the figure while the modeling of the downstream classification tasks and the evaluation on the validation and test sets are shown on the right half of the figures. Based on the standard ICP method, the training dataset is partitioned into proper training set and calibration set. The proper training set is used to represent the noisy training data and the calibration set represents the well-curated dataset. Potential wrongly labeled data and outliers in the proper training set are detected and corrected based on the P-value calibrated on the nonconformity measure distribution on the calibration set. The cleaned training set is then used to train classifiers for downstream classification tasks and compared against the baseline.}
\label{pipeline}
\end{figure*}

\begin{itemize}
    \item Wrongly labeled data detection: if the original label of a training data point has a P-value smaller than another possible label by a large margin (abbreviated as "detection threshold" in the following sections: 0.8/0.5/0.2), the training data point is deemed as wrongly labeled because there is a label that shows better conformity evaluated by the well-curated calibration set;
    \item Wrongly labeled data correction: the label with the highest P-value will be used to replace the original label of the training data point;
    \item Outlier detection: if all of the possible labels are with P-values smaller than 0.1, the training data point is deemed as an outlier since all labels show low conformity to the well-curated calibration set;
    \item Outlier correction: the training data point is removed.
\end{itemize}

Upon detecting the wrongly labeled data and outliers, corrections are made accordingly. The cleaned proper training set is then combined with the calibration set to be the cleaned training set and used for the downstream prediction tasks (right half section of Fig. \ref{pipeline}). For the downstream tasks, we used the linear discriminant analysis (LDA) and logistic regression (LR) as two representative classifiers. The classifier hyperparameters are fixed with the default values in scikit-learn packages, because this study focuses on investigating the influence of the training method process and whether cleaner training data can lead to better classification performances. To investigate how the thresholds to detect the wrongly labeled data affect the performance, we did experiments on three detection thresholds: 0.2, 0.5, and 0.8, and reported all of the results.

For the DILI prediction task, the cleaning method is shown exactly as in Fig. \ref{pipeline}. For the COVID-19 prediction task, before any classification modeling (both in the training of CPSC and the downstream LDA/LR classifier), the training data have been augmented with synthetic minority oversampling technique (SMOTE) to up-sample the positive cases to balance the two classes \cite{xuzhan2021ai}. Additionally, for the COVID-19 prediction task and TCGA breast cancer subtype prediction task, Lasso feature selection was performed before any classification modeling to deal with the high feature dimensionality. 

The hyperparameters of the CPSC ($\Delta$ and $T$), which is the core algorithm of the cleaning process, as well as the strength of L1 penalty in Lasso feature selection ($C$), were tuned based on the performance on the validation dataset. For the three tasks, the metrics used to feedback the hyperparameter tuning were: classification accuracy for the DILI dataset, the sum of AUROC and AUPRC for the COVID-19 dataset, the sum of classification accuracy and macro-average F1 score for the TCGA dataset. The range of the hyperparameters was: 1) in CPSC: $\Delta$: [0, 0.1, 0.2, 0.3], $T$: [1, 10, 100]; 2) in Lasso feature selection: $C$: [0.05, 0.1, 1, 10]. 

To simulate different levels of noises in the training data in the real-world applications, we artificially permuted 0 to 80\% training labels in the proper training set (increment: 10\%). As a result, there were a total of 54 scenarios for each classification task: 3 different detection thresholds, 9 different levels of noises, and 2 different classifiers. 48 scenarios were with the known labeling noises caused by manual label permutation and 6 scenarios were with the original training data.

\subsection{Model performance evaluation}
To evaluate the performance of the training data cleaning method, we investigate both the effectiveness and the cleaning processes of the method. 
The effectiveness of the training data cleaning method is quantified by the improvement of classification metrics on the three datasets. On the DILI literature classification task, the classification accuracy was used as the metric considering the balanced classes in the dataset; on the COVID-19 ICU admission prediction task, the AUROC and AUPRC were used as the metric due to the imbalanced dataset with much fewer positive cases; on the TCGA breast cancer subtype prediction task, the multi-class classification accuracy and macro-average F1 score were used as the metric.

To analyze how the training data cleaning method works and investigate the dynamics of the method, the cleaning processes were visualized. We counted and plotted the total number of corrections for wrongly labeled data detected by the method, the number of wrongly labeled data for each class, as well as the number of outliers detected by the method. Additionally, using the well-curated DILI dataset as an example, we investigate the correctness of the cleaning processes by counting and plotting the total number of wrongly labeled data detected (same as the total number of corrections), the total number of truly wrongly labeled data, the number of wrongly labeled data after correction. How the cleaning processes variate with the detection thresholds for wrongly labeled data was also investigated.

Finally, we investigate how the cleaning processes variate with the hyperparameters in CPSC ($\Delta$ and $T$), which is the core algorithm of the training data cleaning method and the results are shown in the supplementary materials.

\subsection{Statistical tests}
To test the robustness of experiments, we performed random dataset partitions for 30 times and perform each of the 54 scenarios for 30 repetitive experiments. The mean value and 95\% confidence interval (CI) of the classification performance metrics will be reported in the result section. Paired t-tests were performed to test the statistical significance in the metrics with/without training data cleaning.

\section{Results}

\subsection{Effectiveness of training data cleaning on the drug-induced liver injury literature filtering task}
\begin{figure*}[!htp]
\centering
\includegraphics[width=0.9\linewidth]{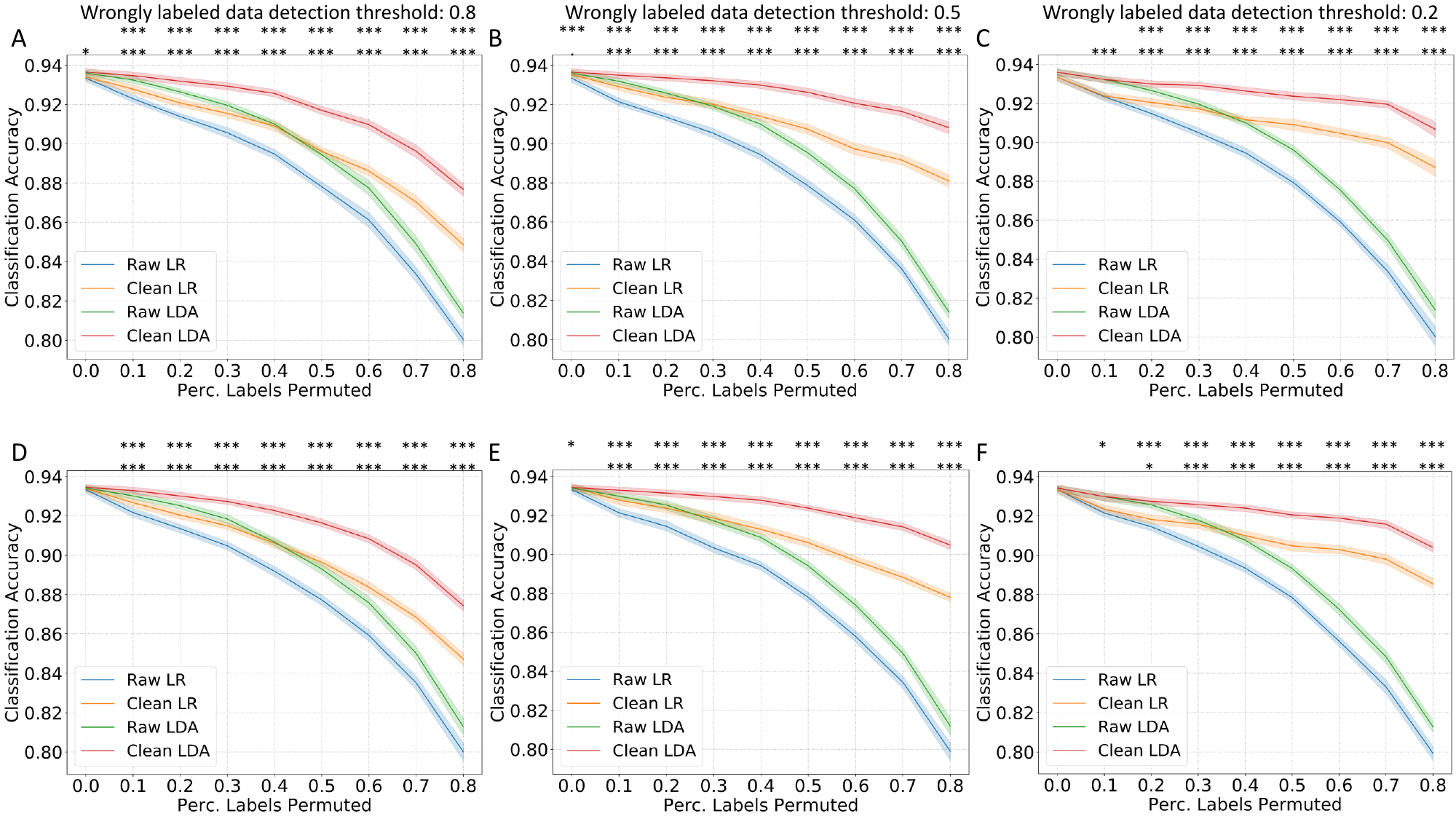}
\caption{The model accuracy improvement with training data cleaning in DILI literature classification task based on the S2V embeddings under different percentages of training data label permutation. The classification accuracy on the validation set (A-C) and on the test set (D-F) with a wrongly labeled data detection threshold of 0.8 (A,D), 0.5 (B,E) and 0.2 (C,F). The mean and 95\% confidence intervals are shown. Statistically significant improvement in accuracy has been marked as follows .: $p<0.1$, *: $p<0.05$, **: $p<0.01$, ***: $p<0.001$; first row: LR models, second row: LDA models.}
\label{DILI S2V}
\end{figure*}

\begin{figure*}[!htp]
\centering
\includegraphics[width=0.9\linewidth]{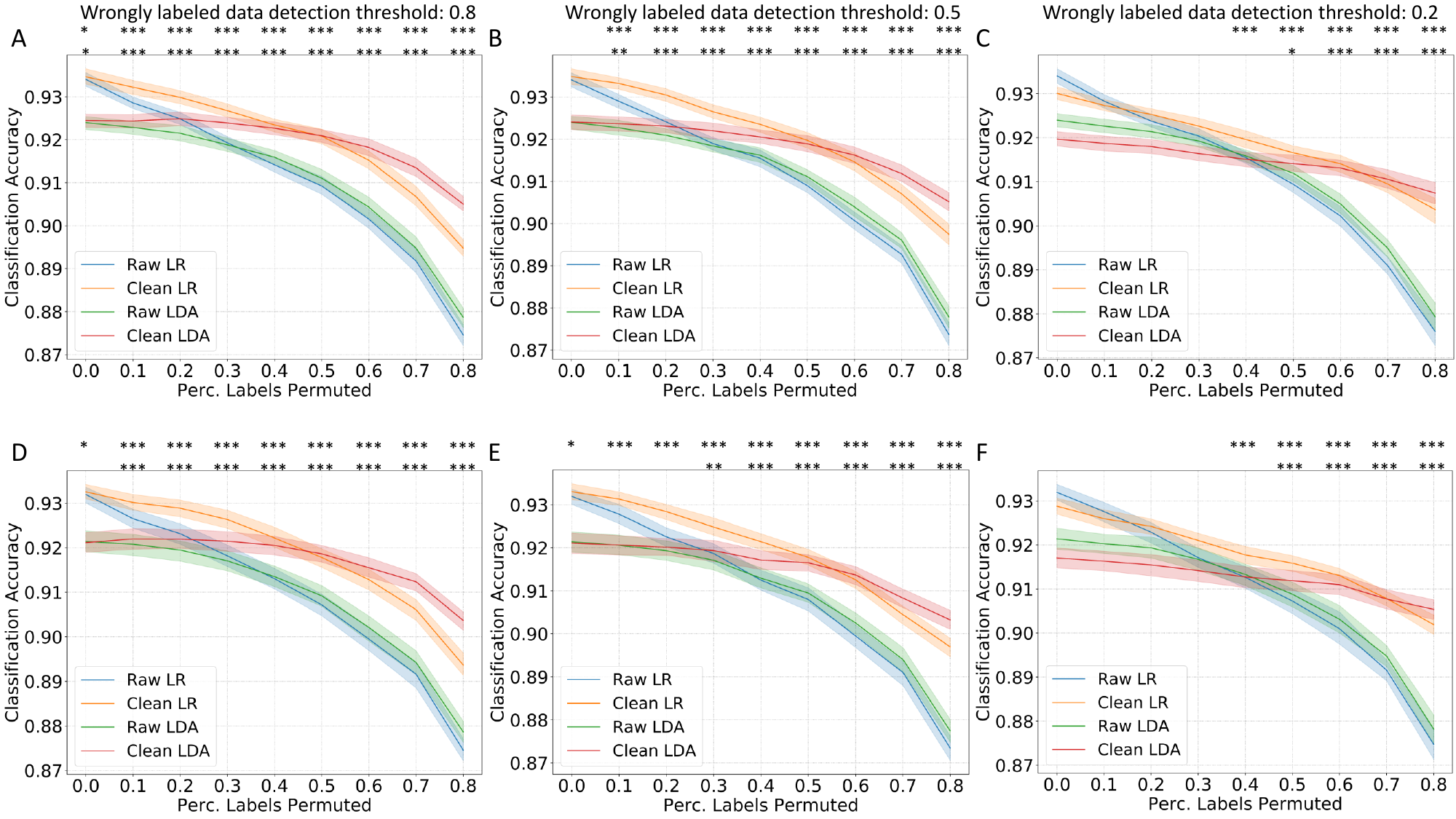}
\caption{The model accuracy improvement with training data cleaning in DILI literature classification task based on the W2V embeddings under different percentages of training data label permutation. The classification accuracy on the validation set (A-C) and on the test set (D-F) with a wrongly labeled data detection threshold of 0.8 (A,D), 0.5 (B,E) and 0.2 (C,F). The mean and 95\% confidence intervals are shown. The statistically significant improvement in accuracy has been marked as follows: .: $p<0.1$, *: $p<0.05$, **: $p<0.01$, ***: $p<0.001$; first row: LR models, second row: LDA models.}
\label{DILI W2V}
\end{figure*}

The effectiveness of inductive conformal prediction in training data cleaning was first evaluated on the DILI classification task: to predict whether a publication has drug-induced-liver-injury contents or not based on the free-text title and abstract. For this task, we considered two types of word embeddings, and the results are respectively shown in Fig. \ref{DILI S2V} (S2V embeddings) and \ref{DILI W2V} (W2V embeddings). Each type of text embedding was tested in 54 scenarios, encompassing two classifiers (LR, LDA), three detection thresholds of wrongly labeled data (0.8, 0.5, 0.2) and nine levels of training label permutation from 0 to 80\% (increment of 10\%). As the percentage of permuted training labels increased, the classification accuracy decreased. In scenarios with no permuted training labels (six scenarios in total), the cleaning process did not yield significant improvements. This outcome can be attributed to the DILI dataset being well curated by FDA experts, suggesting that the original training labels contained minimal noise requiring correction through the cleaning method. 

With the S2V embeddings, when the percentage of labels permuted is larger than zero, with the conformal-prediction-based training data cleaning method, the classification accuracy for both LR and LDA models on the test is statistically significantly better in most scenarios on the test set ($p<0.05$ on 47 out of 48 scenarios with permuted training labels). A higher detection threshold for the wrongly labeled data (a stricter strategy to correct the wrongly labeled data from the perspective of the conformal predictor) also leads to more evident accuracy improvement when the noise in the training data is low (percentage of labels permuted lower than 0.3), while a lower detection threshold can lead to more accuracy improvement when the training labels become noisier (percentage of labels permuted larger than 0.4). The largest improvement in test accuracy (in the term of absolute value of accuracy increment) was in the scenario with 80\% labels permuted and a detection threshold of 0.2 with the LDA classifier: from 0.8120 to 0.9048 (11.4\%).

With the W2V embeddings, similar results were shown: the accuracy was significantly better in most scenarios on the test set ($p<0.05$ on 39 out of 48 scenarios with permuted training labels). The effect of the detection threshold for the wrongly labeled data is similar to that shown based on the S2V embedding. What is more evident is that with a detection threshold of 0.2, the cleaning method cannot guarantee model accuracy improvement when the training data is less noisy (i.e., when the percentage of labels permuted is lower than 0.4). The largest improvement in test accuracy (in the term of the absolute value of accuracy increment) was in the scenario with 80\% labels permuted and a detection threshold of 0.2 with the LR classifier: from 0.8747 to 0.9019 (3.1\%).

\subsection{Effectiveness of training data cleaning on the COVID-19 ICU admission prediction task}
\begin{figure}[!htp]
\centering
\includegraphics[width=\linewidth]{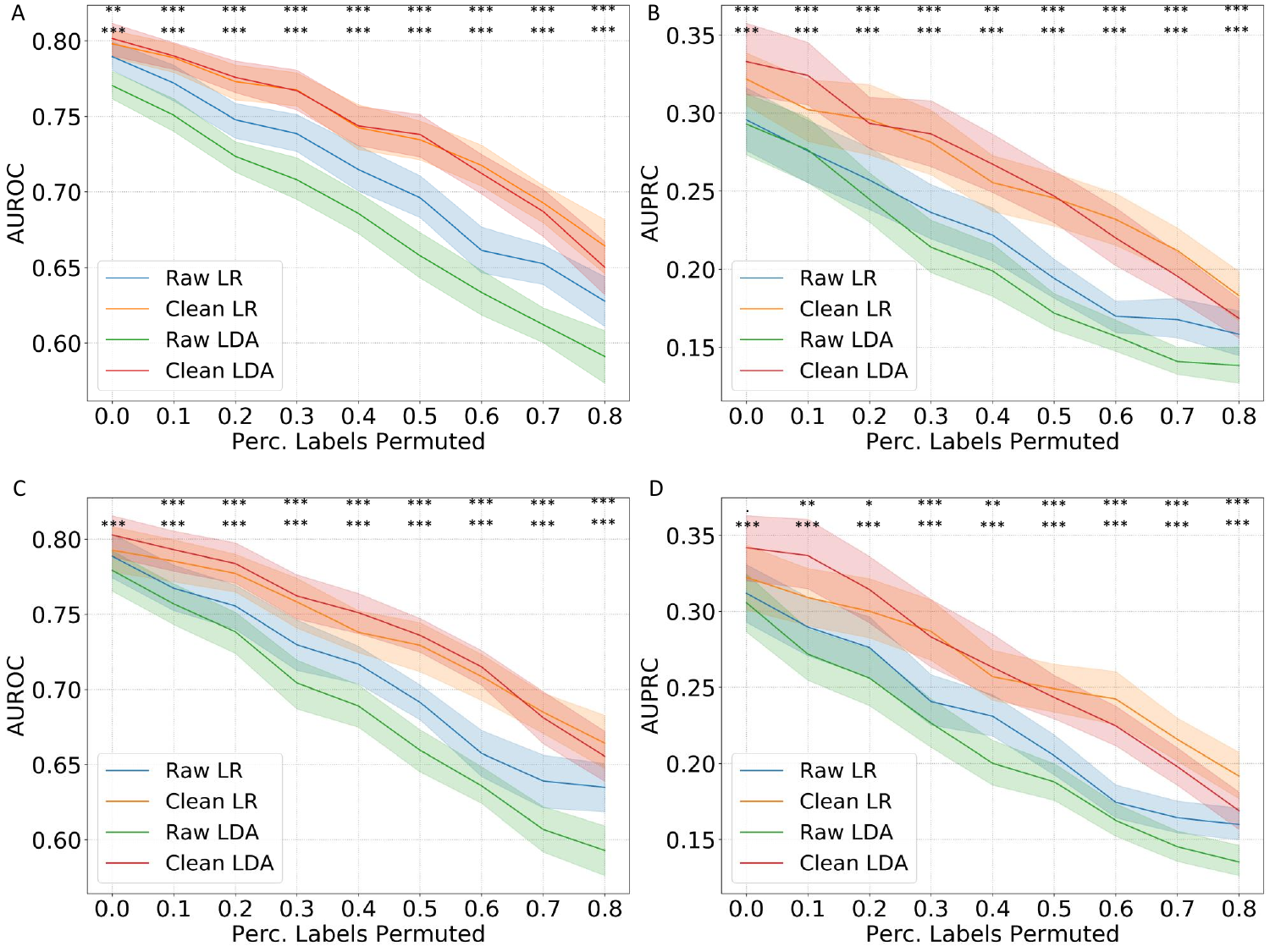}
\caption{The model performance in AUROC and AUPRC with training data cleaning in COVID-19 patient ICU admission prediction task under different percentages of training data label permutation. The AUROC (A) and AUPRC (B) on the validation set, and the AUROC (C) and AUPRC (D) on the test set with a wrongly labeled data detection threshold of 0.8. The mean and 95\% confidence intervals are shown. The statistically significant improvement in accuracy has been marked as follows: .: $p<0.1$, *: $p<0.05$, **: $p<0.01$, ***: $p<0.001$; first row: LR models, second row: LDA models.}
\label{COVID19 0.8}
\end{figure}

\begin{figure}[!htp]
\centering
\includegraphics[width=\linewidth]{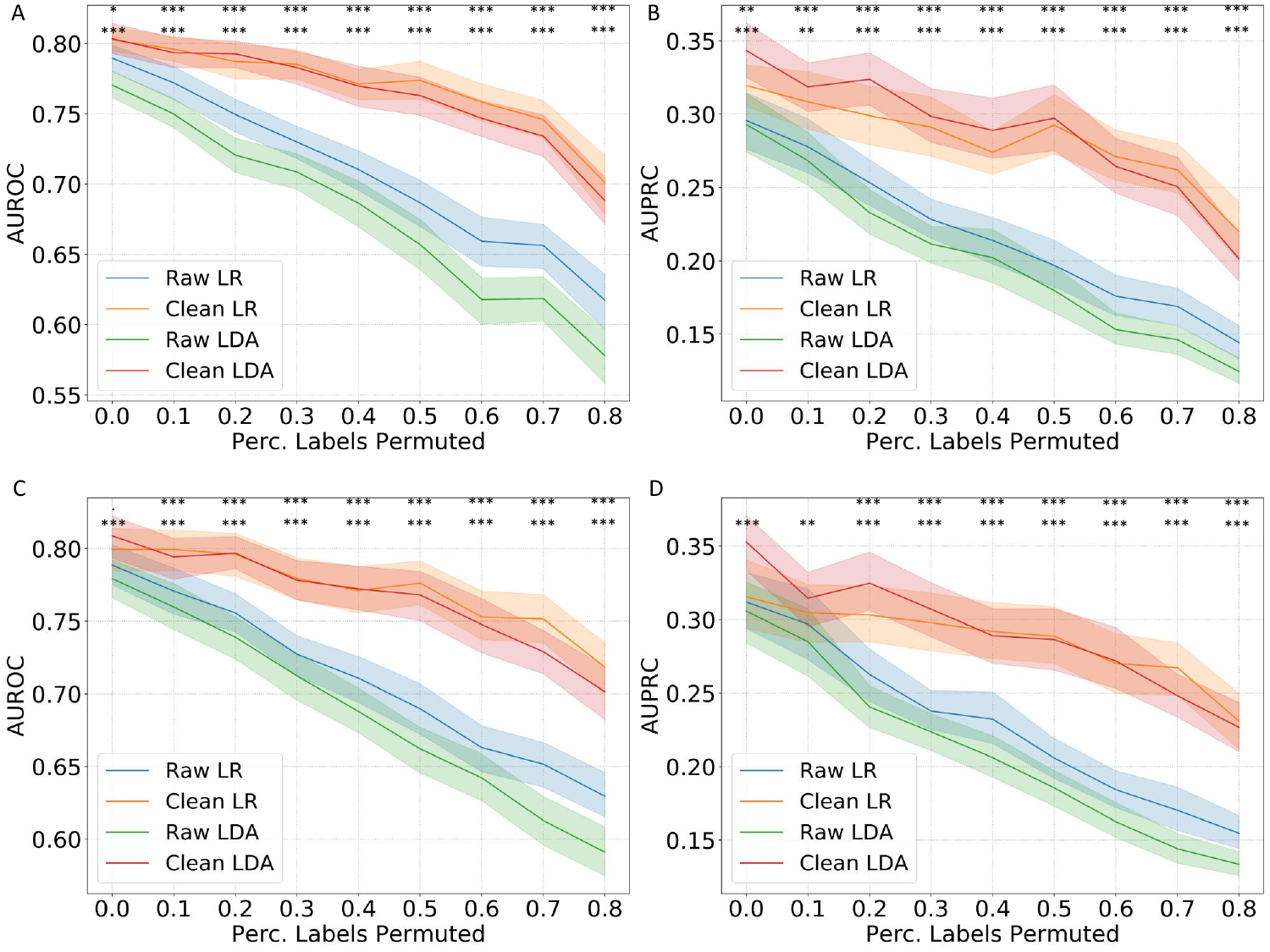}
\caption{The model performance in AUROC and AUPRC with training data cleaning in COVID-19 patient ICU admission prediction task under different percentages of training data label permutation. The AUROC (A) and AUPRC (B) on the validation set, and the AUROC (C) and AUPRC (D) on the test set with a wrongly labeled data detection threshold of 0.5. The mean and 95\% confidence intervals are shown. The statistically significant improvement in accuracy has been marked as follows: .: $p<0.1$, *: $p<0.05$, **: $p<0.01$, ***: $p<0.001$; first row: LR models, second row: LDA models.}
\label{COVID19 0.5}
\end{figure}

In the COVID-19 ICU admission prediction task: to predict whether a COVID-19 patient admitted to the general ward will be admitted to ICU based on the fusion of radiomics data, clinical data and laboratory data, the model performance in AUROC and AUPRC are shown in Fig. \ref{COVID19 0.8} and \ref{COVID19 0.5} (as examples with detection thresholds of 0.8 and 0.5 for wrongly labeled data, the results on the detection threshold of 0.2 are shown in the supplementary materials). Different from the well-curated DILI dataset, even without artificial label permutation, the original proper training dataset in the COVID-19 task may contain noise after the SMOTE up-sampling for the minority data. The results with a detection threshold of 0.8 (Fig. \ref{COVID19 0.8}) show that even without any artificial label permutation (the percentage of labels permuted is 0), the AUROC and AUPRC can be significantly improved  ($p<0.001$ for LDA on both validation and test sets, $p<0.01$ for LR on the validation set). With the increasing percentage of labels permuted, the conformal-prediction-based training data cleaning method shows its effectiveness in improving the AUROC and AUPRC on all scenarios with manual label pollution (n = 48) on the test set ($p<0.05$ for both AUROC and AUPRC in all scenarios with label permutations). Similar results were shown with a detection threshold of 0.5 (Fig. \ref{COVID19 0.5}): significantly improved AUROC and AUPRC with LDA models without any label permutation ($p<0.001$), and significantly improved AUROC and AUPRC with label permutation ($p<0.05$ in all scenarios except for the AUPRC with LR with 10\% label permutation) on the test set. The largest improvement in test AUROC (in the term of absolute value of increment) was in the scenario with 80\% labels permuted and a detection threshold of 0.2 with the LDA classifier: from 0.5967 to 0.7389 (23.8\%) while the largest improvement in test AUPRC was in the scenario with 50\% labels permuted and a detection threshold of 0.2 with the LDA classifier: from 0.1829 to 0.3106 (69.8\%).

\subsection{Effectiveness of training data cleaning on the breast cancer subtype prediction task}

\begin{figure}[!htp]
\centering
\includegraphics[width=\linewidth]{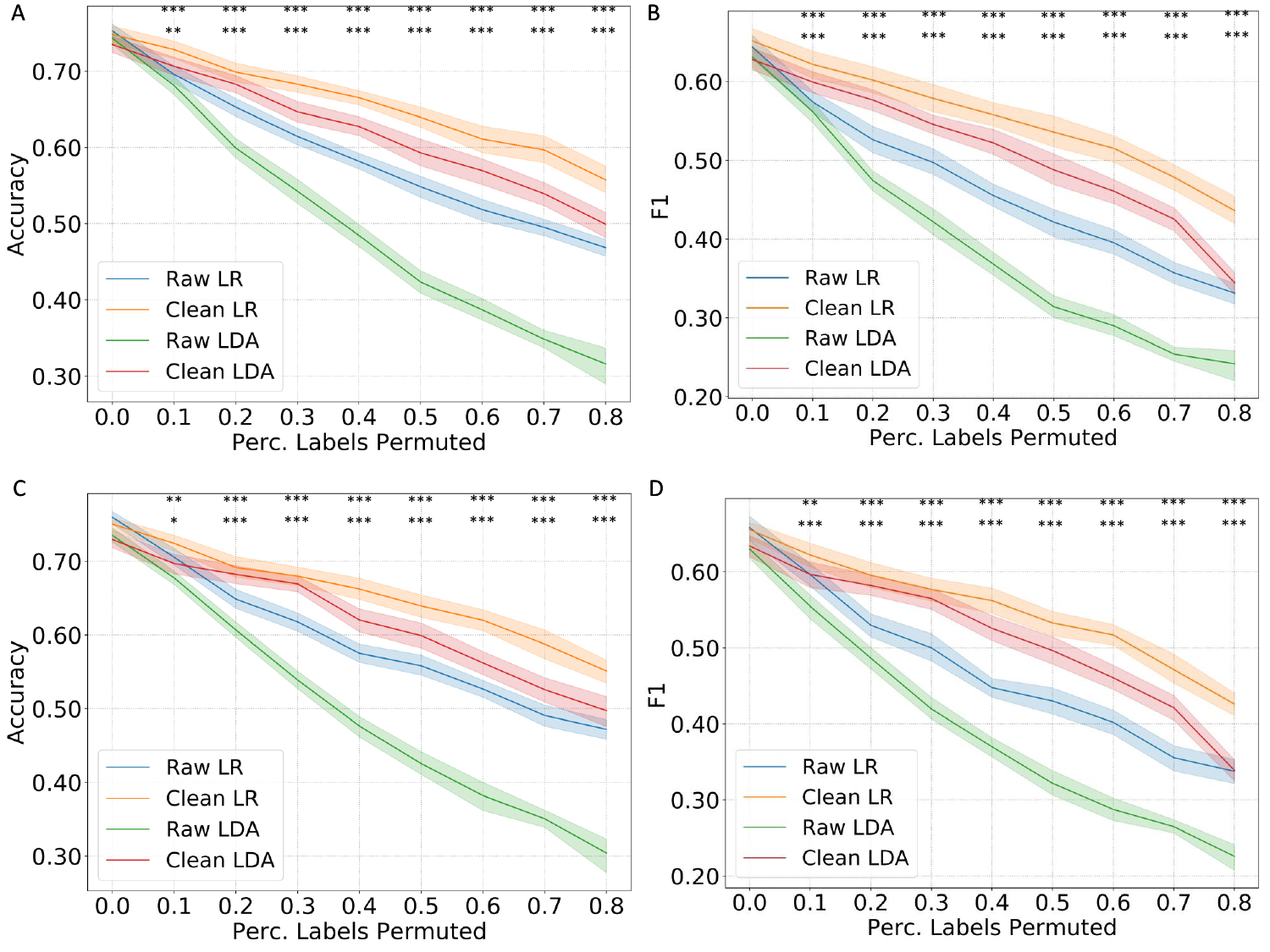}
\caption{The model performance in accuracy and F1 score with training data cleaning in the breast cancer subtype prediction task under different percentages of training data label permutation. The classification accuracy (A) and macro-averaged F1 score (B) on the validation set, and the classification accuracy (C) and macro-averaged F1 score (D) on the test set with a wrongly labeled data detection threshold of 0.5. The mean and 95\% confidence intervals are shown. The statistically significant improvement in accuracy has been marked as follows: .: $p<0.1$, *: $p<0.05$, **: $p<0.01$, ***: $p<0.001$; first row: LR models, second row: LDA models.}
\label{RNAseq 0.5}
\end{figure}

We next evaluated the effectiveness of reliability-based training data cleaning method in classifying the molecular subtypes of breast cancer using RNA-seq data of the TCGA cohort (n = 1,096 patients)  \cite{cancer2012comprehensive}.  We considered five subtypes in our analysis, namely LumA (n = 569), LumB (n = 219), Basal (n = 187), HER2 (n = 82), and normal-like (n = 39). The accuracy and macro-averaged F1 score are shown in Fig. \ref{RNAseq 0.5} (as an example with detection thresholds of 0.5 for wrongly labeled data; the results on the detection thresholds of 0.8 and 0.2 are shown in the supplementary materials). In the absence of noises in training labels, the cleaning method did not yield a significant improvement in performance, in terms of both classification accuracy and macro-average F1 score. However, when dealing with scenarios containing noisy training labels (n = 48), the data cleaning method leads to a significant improvement in classification accuracy ($p<0.05$ for 47 out of 48 scenarios) and macro-averaged F1 score ($p<0.05$ for 47 out of 48 scenarios). Similar to the findings from previous tasks, we observed that using lower detection thresholds (less strict rules to detect wrongly labeled data) resulted in better improvements in classification accuracy and macro-averaged F1 score, especially in highly noisy scenarios where over 40\% training labels were permuted. The largest improvement in test accuracy (as quantified by the absolute value of increment) was observed when 70\% of labels were permuted, and the detection threshold was set to 0.2, using the LDA classifier.  In this scenario, accuracy increased by 74.6\% (from 0.3508 to 0.6128), while the macro-average F1 score improved by 89.0\% (from 0.2672 to 0.5049). Overall, these results demonstrate the effectiveness of our data-cleaning method in improving the classification performance on the TCGA RNA-seq dataset.

\subsection{Analyses of the cleaning processes based on inductive conformal prediction}
\begin{figure*}[!htp]
\centering
\includegraphics[width=0.9\linewidth]{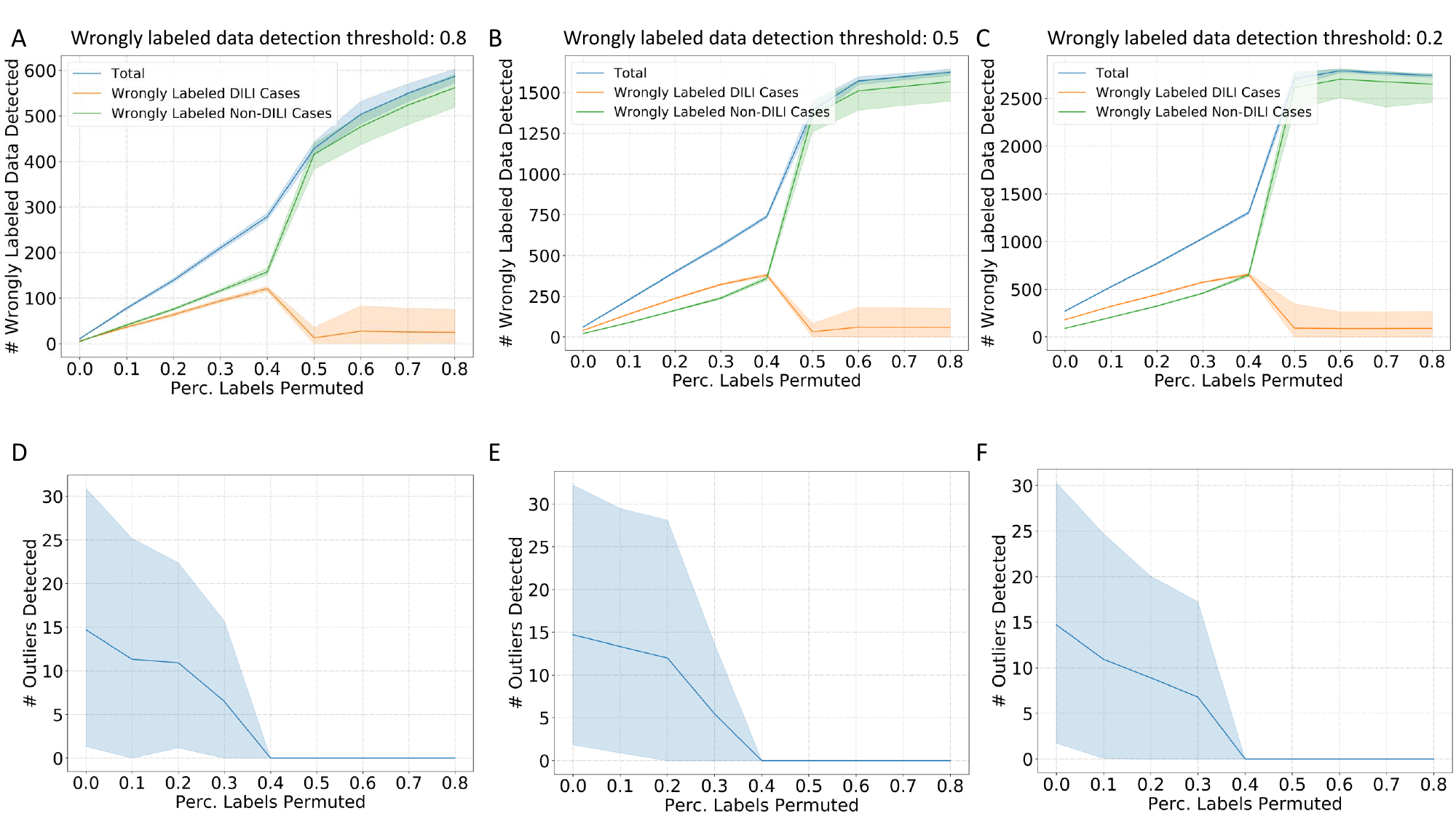}
\caption{The number of wrong labels and outliers detected under different percentages of training data label permutation in DILI literature prediction task with W2V embeddings. The number of wrongly labeled data (A-C) and outliers (D-F) under different detection thresholds of wrongly labeled data: 0.8 (A,D), 0.5 (B,E), 0.2 (C,F). The cleaning process visualization is based on W2V embeddings and fixed hyperparameters for the conformal predictor.}
\label{DILI W2V Cleaning Process}
\end{figure*}

\begin{figure*}[!htp]
\centering
\includegraphics[width=0.9\linewidth]{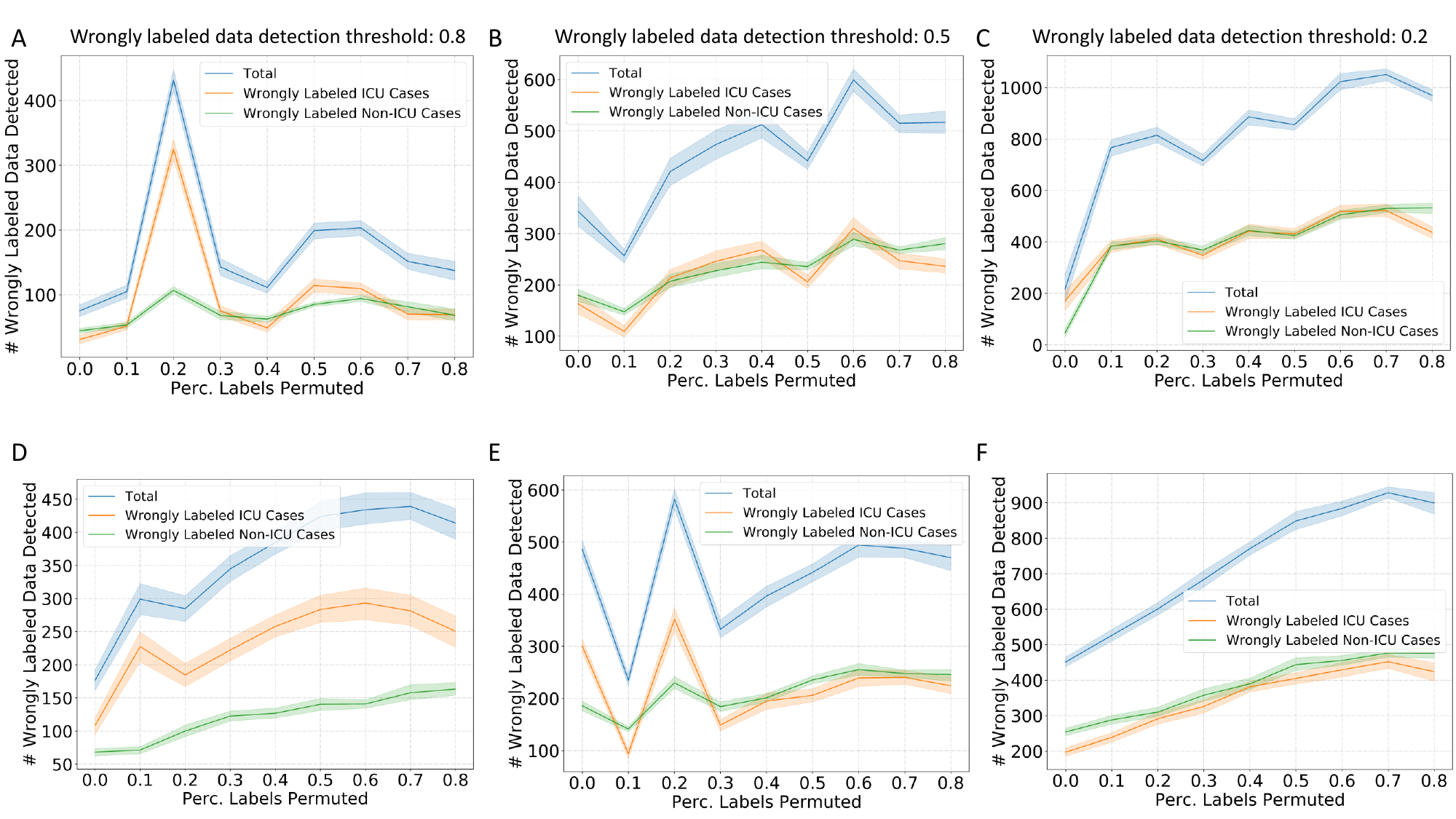}
\caption{The number of wrong labels detected under different percentages of training data label permutation in COVID-19 patient ICU admission prediction task. The number of wrongly labeled data based on LR models (A-C) and LDA models (D-F) under different detection thresholds of wrongly labeled data: 0.8 (A,D), 0.5 (B,E), 0.2 (C,F). The cleaning process visualization is based on optimized hyperparameters for the conformal predictor tuned on the validation dataset for each classifier and each percentage of labels permuted.}
\label{COVID Cleaning Process}
\end{figure*}

\begin{figure*}[!htp]
\centering
\includegraphics[width=0.9\linewidth]{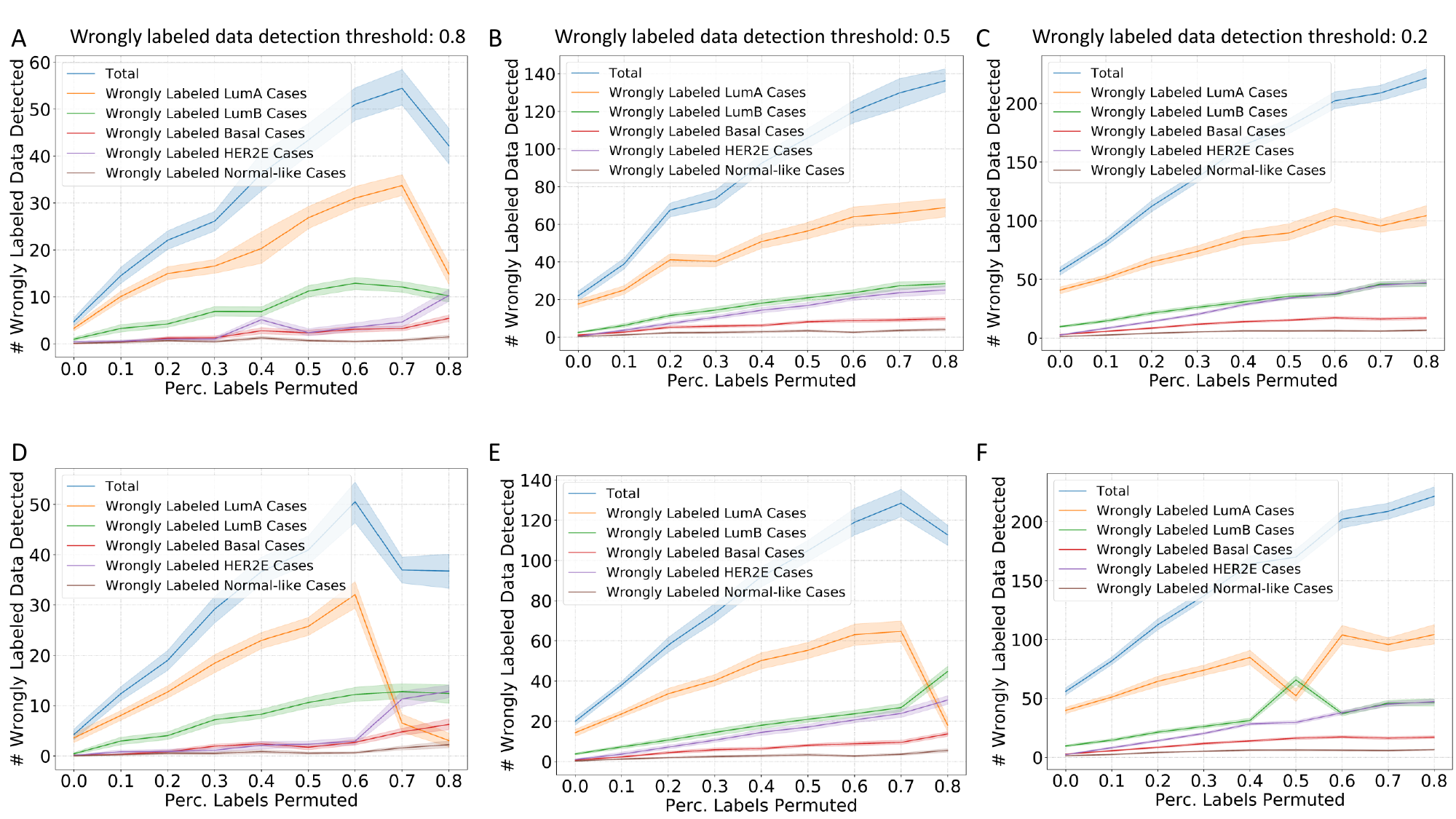}
\caption{The number of wrong labels detected under different percentages of training data label permutation in TCGA breast cancer subtype prediction task. The number of wrongly labeled data based on LR models (A-C) and LDA models (D-F) under different detection thresholds of wrongly labeled data: 0.8 (A,D), 0.5 (B,E), 0.2 (C,F). The cleaning process visualization is based on optimized hyperparameters for the conformal predictor tuned on the validation dataset for each classifier and each percentage of labels permuted.}
\label{TCGA Cleaning Process}
\end{figure*}

To show the detailed processes of the training data cleaning method, the number of wrongly labeled data and outliers are calculated and visualized in Fig. \ref{DILI W2V Cleaning Process}, \ref{COVID Cleaning Process} and \ref{TCGA Cleaning Process}. Firstly, we used the DILI W2V dataset as an example to showcase how the cleaning process behaves under different detection thresholds for wrongly labeled data when the hyperparameters of the conformal predictor (CPSC) are fixed (Fig. \ref{DILI W2V Cleaning Process}, $\Delta$ = 0.3 and $T$ = 100). As the percentage of labels permuted increases and the proper training data becomes noisier, the number of wrongly labeled data detected by the method increases while it is harder to detect outliers which indicates that the models are less confident to ascertain outliers. Additionally, as the detection thresholds are set lower, more wrongly labeled data are detected and more label corrections are made by the training data cleaning method. Meanwhile, as the training data grows to be noisier, the detection can be biased towards the wrongly labeled positive cases (DILI-related publications). 

\begin{figure*}[!htp]
\centering
\includegraphics[width=0.9\linewidth]{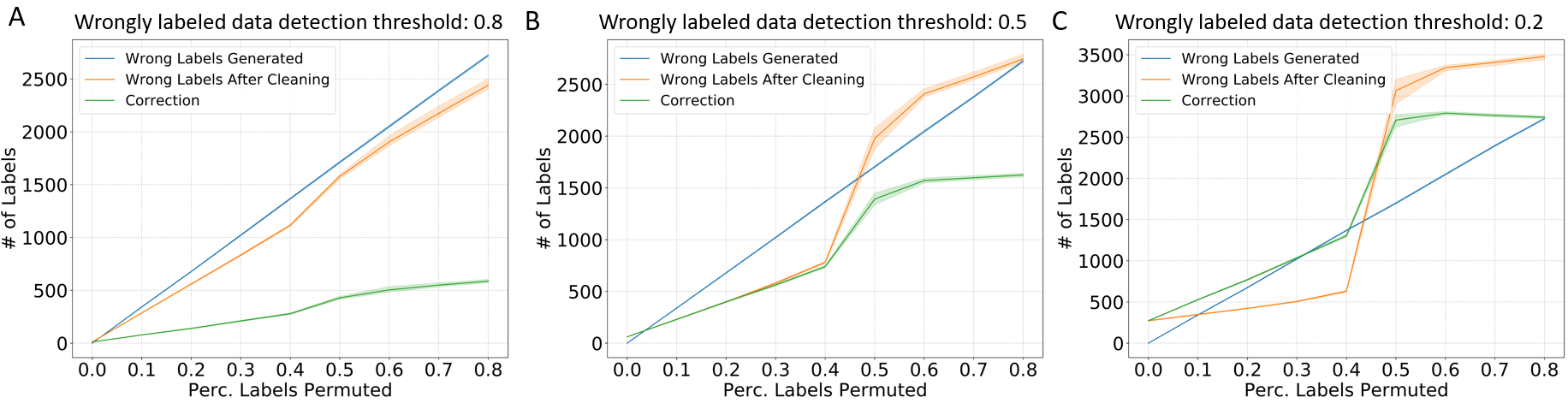}
\caption{The number of ground-truth wrong labels before and after training data cleaning under different percentages of training data label permutation in DILI literature prediction task with W2V embeddings. The number of wrongly labeled data before/after training data cleaning and the number of corrections made under different detection thresholds of wrongly labeled data: 0.8 (A), 0.5 (B), 0.2 (C). The cleaning process visualization is based on W2V embeddings and fixed hyperparameters for the conformal predictor.}
\label{DILI W2V Cleaning Correctness}
\end{figure*}

Because the DILI data set is well-curated with high-fidelity labels, we were also able to evaluate the correctness of the corrections made by the training data cleaning method by directly comparing the cleaned training data set and the original noisy training data set. Here, we visualized the number of ground-truth wrong labels before and after the training data cleaning processes, as well as the total number of corrections made by the data cleaning method in Fig. \ref{DILI W2V Cleaning Correctness}. When the models are stricter in detecting wrongly labeled data (when the detection threshold is set to 0.8), the models are effective in reducing the number of wrongly labeled data after making the corrections. When the detection thresholds are set lower (0.5 and 0.2), when the percentage of labels permuted is lower than 0.5, the corrections are effective in reducing the number of wrongly labeled data. However, as the percentage goes above 0.5 and the training data contain more noise than signals, the data cleaning method can lead to over-correction: after the cleaning process, the number of wrongly labeled data can be even higher.

Additionally, we showed the cleaning process in the COVID-19 patient ICU admission prediction task in Fig. \ref{COVID Cleaning Process}. Different from Fig. \ref{DILI W2V Cleaning Process}, we visualized the cleaning process after the CPSC has been optimized via hyperparameter tuning, respectively based on LR and LDA classifiers. Similar to the results on DILI dataset, a lower detection threshold leads to more wrongly labeled data detected and more corrections, and as the training data become noisier, generally, more corrections for the wrongly labeled data are observed. However, in this task, the wrongly labeled positive cases (patients admitted to the ICU) and wrongly labeled negative cases (patients staying in the general ward) are more balanced. The hyperparameter tuning process leads to certain spikes of wrongly labeled data detected at certain noise levels. In addition, the models are not confident in telling outliers in this task. 

For the breast cancer subtype prediction task, similar observations are shown in Fig. \ref{TCGA Cleaning Process}: lower detection thresholds lead to more wrongly labeled data being detected and curated; the number of wrongly labeled data detected for each category correspond with the prevalence of each subtype.

\section{Discussion}

To address the scenario with the challenge of collecting well-curated labels and the hardship in accurately labeling biomedical data with high-fidelity for supervised learning, this study proposes a reliability-based training data cleaning method based on inductive conformal prediction. With a small portion of well-curated training data (in the calibration set of the inductive conformal prediction framework), our proposed method leverages the reference distribution of nonconformity measure on the calibration set to calibrate the conformity of the noisy labels in the large portion of training data with unknown label quality (in the proper training data of the inductive conformal prediction framework). By simulating scenarios with different levels of noise in the proper training data by manually permuting the training labels, we validated the reliability-based training data cleaning method on three biomedical machine learning tasks representing different modalities: the drug-induced-liver-injury literature filtering challenge (a natural language processing task), the COVID-19 patient ICU admission prediction task (a radiomics and electronic health records task) and the breast cancer subtype prediction task (an RNA-seq data mining task). The training data cleaning method showed its effectiveness on the majority of the simulated scenarios on all three tasks where the classification performance with two basic LDA and LR classifiers was significantly improved after the data cleaning process. The visualization of the cleaning processes also showed that the model was effective in detecting wrongly labeled data based on the reliability quantified by the inductive conformal prediction framework. Our results demonstrate that this method can be used in a broad range of multi-modal biomedical classification applications to help improve the classification performance without the requirement of large quantities of well-curated labeled training dataset.

The novelty of this study is the proposal of a reliability-based training data cleaning method based on inductive conformal prediction, which enables users to leverage large quantities of noisy labels with the requirement of only a small portion of well-curated training data. Different from the traditional semi-supervised learning which typically requires strong assumptions of the data distributions for the unlabeled data and labeled data \cite{chapelle2009semi}, the inductive conformal prediction is based on a weaker assumption of independent and identical distribution. Moreover, this method leverages both well-curated labels and noisy labels with uncertainty. This idea of this work was inspired by the semi-supervised, reliability-based training data augmentation work based on conformal prediction previously proposed \cite{liu2022cpsc,liu2021boost,wang2022unsupervised}. In previous work, researchers first attach pseudo-labels to the unlabeled data and then leverage the prediction reliability of the pseudo-labels quantified by conformal predictors, to filter these pseudo-labeled data. On the classification tasks, even with domain drifts between the training and test datasets, the reliability-based unlabeled data augmentation framework showed significantly better performances when compared with multiple baseline models: fully supervised learning benchmark, as well as other semi-supervised learning (i.e., label propagation \cite{bengio200611} and label spreading \cite{chen2017combining}) and data augmentation frameworks \cite{liu2021boost}. These studies follow the idea of reliability-based unlabeled data augmentation, which inspired us to think in the opposite direction with the idea of reliability-based training data reduction and trimming: as we can leverage the reliability quantified by the conformal predictors to filter and add the unlabeled data to benefit the classification modeling process, we should also be able to leverage the reliability to remove the wrongly labeled data in a noisy training set that may confuse the classification modeling process. The rich information given by the nonconformity measure of the conformal predictors enables us to correct the noisy labels in the training data and detect the wrongly labeled data. After implementing this idea, we found that the reliability-based training data cleaning method works well in diverse multi-modal biomedical data sets and classification tasks.

The mechanism of the reliability-based training data cleaning method is worthy of further discussion. The method quantifies how well a combination of a training sample's feature with every possible label conforms to the reference distribution of the calibration set. By leveraging and calibrating the nonconformity measure distribution on a small portion of well-curated training data (calibration set), the method can detect whether a training sample may be wrongly labeled after trying out all possible labels that are attached to a training sample's feature. As our method strictly follows the framework of inductive conformal prediction, the training data after cleaning tends not to overfit as both the noisy training data and the clean calibration data set are used at two separate stages: the noisy proper training data were used to fit the conformal predictor while the clean calibration set was used for calibration. The bulk size of the proper training data helps the conformal predictor to learn a general yet fuzzy mapping from the noisy data to the classes while the calibration process better clean the training data by filtering out the labels that are extremely unlikely. Instead of using the same dataset to train conformal predictor and calibrate the reliability of predictions, which have been used in some of the previous studies (the non-inductive conformal prediction framework) \cite{zhan2022filter,zhan2020electronic,liu2021boost}, the relatively better independence of these two sets based on the inductive conformal prediction framework helps the training data avoid biasing towards the small portion of the calibration set by dissociating the two processes in inductive conformal prediction. As a result, the cleaned training data (the combination of the cleaned proper training set and the calibration set) leads to better classification performances.

Another adaptive feature of the training data cleaning method is the freedom for users to choose the detection threshold of wrongly labeled data, more specifically how much another label's P-value needs to be larger than the current label's P-value to detect a wrongly labeled data point. We have observed that for less noisy training sets (i.e. the percentage of labels permuted is lower than 50\%), a higher detection threshold (e.g., 0.8) is more likely to lead to significant improvement in classification performance. On the contrary, for highly noisy proper training sets (percentage of labels permuted over 50\%), a lower detection threshold (e.g., 0.2) can lead to better improvement in classification performance. We hypothesize that once the training set gets noisier, less classification information is conveyed in the proper training dataset. Therefore, the CPSC model can be less confident in telling wrongly labeled data. Lowering the detection threshold, in this case, enables more wrongly labeled data to be detected to counteract low confidence in the CPSC model. Therefore, if future users of our method have knowledge of the noise level in the training dataset (e.g., a rough idea of how many labels may be wrongly labeled in a particular dataset), they can choose a high or low detection threshold to help optimize the performance of the training data cleaning method.

Some observations in the results are also worth further discussion. Firstly, on the DILI task we have observed as the training data grows to be noisier, the detection can bias towards the wrongly labeled positive cases (DILI-related publications) (Fig. \ref{DILI W2V Cleaning Process}). We assume this is because, in this task, the negatives are by default where a publication is absent of DILI information. This means that the negative samples can be highly heterogeneous: a diverse range of publications associated with vaccine development, optogenetics, epigenetics, transcriptomics, neurological disorders, etc. can be labeled as DILI-negative papers. The DILI-negative samples may not be necessarily a clearly defined class based on contents but grouped because of the absence of DILI information. On the contrary, the positive cases (i.e., the DILI-related publications) are relatively more homogeneous. Therefore, the model tends to be more confident in telling whether a DILI-related publication is wrongly labeled as a DILI-irrelevant publication. In contrast, for the COVID-19 dataset and the TCGA RNA-seq dataset, the classes are more clearly defined.

Secondly, we have observed over-correction in the DILI task in the visualization of the correctness of the training process of the W2V classification task (Fig. \ref{DILI W2V Cleaning Correctness}): when the percentage of labels permuted is over 50\%, the cleaning can lead to more wrongly labeled data. However, it should be mentioned that although over-correction was observed, the cleaning process still showed effectiveness in improving classification accuracy. We hypothesize that this may potentially be due to that this visualization of cleaning processes is based on fixed hyperparameters with sub-optimal CPSC models, and secondly, potentially these over-corrections may be less influential in the decision boundary determination.

Additionally, although the method was able to detect outliers in the DILI classification task (Fig. \ref{DILI W2V Cleaning Process}, no outliers were detected in the COVID-19 dataset. We hypothesize that due to that the COVID-19 dataset is noisier than the well-curated DILI dataset because of the SMOTE data augmentation and potential wrong labels in the multi-institute data collection process, the models are less confident in judging outliers in this application.

Although this study has proposed an effective training data cleaning method based on inductive conformal prediction, there are limitations: first, conformal predictions are more generally used in classification tasks to quantify prediction reliability. How much the idea of reliability-based training data cleaning can benefit regression tasks remains unknown. The performance of such a training data cleaning framework needs to be developed and validated in regression tasks to further expand the applicability. Secondly, in this study, we only tested one inductive conformal prediction mechanism, namely, the conformal prediction based on shrunken centroids (CPSC). We chose this framework because previous work showed better efficacy and better efficiency in quantifying the reliability when compared with conformal prediction based on k-nearest neighbors (CPKNN), support vector machines (CPSVM), light gradient-boosting machine (CPLGB) and artificial neural networks (CPANN) \cite{liu2022cpsc}. Although the CPSC tends to be more effective and efficient in the reliability quantification process and was more effective in the reliability-based unlabeled data augmentation process, the question whether our training data cleaning method using other base algorithms works better still remains to be tested. Thirdly, the tasks we tested are with large quantities of samples. With over 500 samples, the partition of the calibration set and proper training set leave both sets not too small and this may have enabled us to better leverage the inductive conformal prediction framework. How well the method works on smaller datasets, and whether we need to use one single clean training set to train conformal predictor as well as perform calibration (generally abandon the inductive conformal prediction framework) at the sacrifice of the overfitting risks, need to be further investigated.  

\section{Conclusion}
Collecting well-curated training data has posed a challenge for biomedical machine-learning applications. To address this challenge, a reliability-based training data cleaning method based on inductive conformal prediction has been proposed in this study. With a small portion of well-curated training data, our method leverages the reliability quantified by inductive conformal prediction to detect the wrongly labeled data and outliers in the large portion of noisy-labeled training data. The effectiveness of this method is validated on three multi-modal biomedical machine learning classification tasks: detecting drug-induced-liver-injury literature based on free-text title and abstract, predicting ICU admission of COVID-19 patients based on radiomics and electronic health records, subtyping breast cancer based on RNA-seq data. The method generally leads to significantly improved classification performances under different levels of labeling noises simulated by manual label permutation. The method can be applied to multi-modal biomedical machine learning classification tasks to better use the noisy training data without requiring large quantities of well-curated training data. 

\section{Code Availability}
\url{https://github.com/xzhan96-stf/icp_train_clean}

\ifCLASSOPTIONcaptionsoff
  \newpage
\fi

\bibliographystyle{IEEEtran}
\bibliography{cite}

\begin{thebibliography}{10}
\providecommand{\url}[1]{#1}
\csname url@samestyle\endcsname
\providecommand{\newblock}{\relax}
\providecommand{\bibinfo}[2]{#2}
\providecommand{\BIBentrySTDinterwordspacing}{\spaceskip=0pt\relax}
\providecommand{\BIBentryALTinterwordstretchfactor}{4}
\providecommand{\BIBentryALTinterwordspacing}{\spaceskip=\fontdimen2\font plus
\BIBentryALTinterwordstretchfactor\fontdimen3\font minus
  \fontdimen4\font\relax}
\providecommand{\BIBforeignlanguage}[2]{{%
\expandafter\ifx\csname l@#1\endcsname\relax
\typeout{** WARNING: IEEEtran.bst: No hyphenation pattern has been}%
\typeout{** loaded for the language `#1'. Using the pattern for}%
\typeout{** the default language instead.}%
\else
\language=\csname l@#1\endcsname
\fi
#2}}
\providecommand{\BIBdecl}{\relax}
\BIBdecl

\bibitem{desautels2016prediction}
T.~Desautels, J.~Calvert, J.~Hoffman, M.~Jay, Y.~Kerem, L.~Shieh,
  D.~Shimabukuro, U.~Chettipally, M.~D. Feldman, C.~Barton \emph{et~al.},
  ``Prediction of sepsis in the intensive care unit with minimal electronic
  health record data: a machine learning approach,'' \emph{JMIR medical
  informatics}, vol.~4, no.~3, p. e5909, 2016.

\bibitem{xuzhan2021ai}
Q.~Xu, X.~Zhan, Z.~Zhou, Y.~Li, P.~Xie, S.~Zhang, X.~Li, Y.~Yu, C.~Zhou,
  L.~Zhang \emph{et~al.}, ``Ai-based analysis of ct images for rapid triage of
  covid-19 patients,'' \emph{NPJ digital medicine}, vol.~4, no.~1, pp. 1--11,
  2021.

\bibitem{steyaert2023multimodal}
S.~Steyaert, M.~Pizurica, D.~Nagaraj, P.~Khandelwal, T.~Hernandez-Boussard,
  A.~J. Gentles, and O.~Gevaert, ``Multimodal data fusion for cancer biomarker
  discovery with deep learning,'' \emph{Nature Machine Intelligence}, vol.~5,
  no.~4, pp. 351--362, 2023.

\bibitem{zheng2023spatial}
Y.~Zheng, F.~Carrillo-Perez, M.~Pizurica, D.~H. Heiland, and O.~Gevaert,
  ``Spatial cellular architecture predicts prognosis in glioblastoma,''
  \emph{Nature Communications}, vol.~14, no.~1, p. 4122, 2023.

\bibitem{53_NLP_clinical_texts}
X.~Z. M. H.-D. P.~M. Gevaert, ``Structuring clinical text with ai: old vs. new
  natural language processing techniques evaluated on eight common
  cardiovascular diseases,'' \emph{medRxiv}, 2021.

\bibitem{xu2021advanced}
Q.~Xu, Z.~Sun, X.~Li, C.~Ye, C.~Zhou, L.~Zhang, and G.~Lu, ``Advanced gastric
  cancer: Ct radiomics prediction and early detection of downstaging with
  neoadjuvant chemotherapy,'' \emph{European Radiology}, vol.~31, no.~11, pp.
  8765--8774, 2021.

\bibitem{HuangXu2023distinguishing}
M.~Huang, Q.~Xu, M.~Zhou, X.~Li, W.~Lv, C.~Zhou, R.~Wu, Z.~Zhou, X.~Chen,
  C.~Huang \emph{et~al.}, ``Distinguishing multiple primary lung cancers from
  intrapulmonary metastasis using ct-based radiomics,'' \emph{European Journal
  of Radiology}, vol. 160, p. 110671, 2023.

\bibitem{zhan2022filter}
X.~Zhan, F.~Wang, and O.~Gevaert, ``Reliably filter drug-induced liver injury
  literature with natural language processing and conformal prediction,''
  \emph{IEEE Journal of Biomedical and Health Informatics}, vol.~26, no.~10,
  pp. 5033--5041, 2022.

\bibitem{mccarthy2019}
C.~McCarthy, S.~Murphy, J.~A. Cohen, S.~Rehman, M.~Jones-O’Connor, D.~S.
  Olshan, A.~Singh, M.~Vaduganathan, J.~L. Januzzi, and J.~H. Wasfy,
  ``Misclassification of myocardial injury as myocardial infarction:
  implications for assessing outcomes in value-based programs,'' \emph{JAMA
  cardiology}, vol.~4, no.~5, pp. 460--464, 2019.

\bibitem{chang2016}
T.~E. Chang, J.~H. Lichtman, L.~B. Goldstein, and M.~G. George, ``Accuracy of
  icd-9-cm codes by hospital characteristics and stroke severity: Paul
  coverdell national acute stroke program,'' \emph{Journal of the American
  Heart Association}, vol.~5, no.~6, p. e003056, 2016.

\bibitem{goldstein1998accuracy}
L.~B. Goldstein, ``Accuracy of icd-9-cm coding for the identification of
  patients with acute ischemic stroke: effect of modifier codes,''
  \emph{Stroke}, vol.~29, no.~8, pp. 1602--1604, 1998.

\bibitem{hassan2022supervised}
H.~Hassan, Z.~Ren, C.~Zhou, M.~A. Khan, Y.~Pan, J.~Zhao, and B.~Huang,
  ``Supervised and weakly supervised deep learning models for covid-19 ct
  diagnosis: A systematic review,'' \emph{Computer Methods and Programs in
  Biomedicine}, vol. 218, p. 106731, 2022.

\bibitem{yang2020weakly}
G.~Yang, C.~Wang, J.~Yang, Y.~Chen, L.~Tang, P.~Shao, J.-L. Dillenseger,
  H.~Shu, and L.~Luo, ``Weakly-supervised convolutional neural networks of
  renal tumor segmentation in abdominal cta images,'' \emph{BMC medical
  imaging}, vol.~20, pp. 1--12, 2020.

\bibitem{jain2016weakly}
S.~Jain, T.-T. Kuo, S.~Bhargava, G.~Lin, C.-N. Hsu \emph{et~al.}, ``Weakly
  supervised learning of biomedical information extraction from curated data,''
  in \emph{BMC bioinformatics}, vol.~17, no.~1.\hskip 1em plus 0.5em minus
  0.4em\relax BioMed Central, 2016, pp. 1--12.

\bibitem{prest2011weakly}
A.~Prest, C.~Schmid, and V.~Ferrari, ``Weakly supervised learning of
  interactions between humans and objects,'' \emph{IEEE Transactions on Pattern
  Analysis and Machine Intelligence}, vol.~34, no.~3, pp. 601--614, 2011.

\bibitem{qi2020small}
G.-J. Qi and J.~Luo, ``Small data challenges in big data era: A survey of
  recent progress on unsupervised and semi-supervised methods,'' \emph{IEEE
  Transactions on Pattern Analysis and Machine Intelligence}, vol.~44, no.~4,
  pp. 2168--2187, 2020.

\bibitem{adeli2018semi}
E.~Adeli, K.-H. Thung, L.~An, G.~Wu, F.~Shi, T.~Wang, and D.~Shen,
  ``Semi-supervised discriminative classification robust to sample-outliers and
  feature-noises,'' \emph{IEEE transactions on pattern analysis and machine
  intelligence}, vol.~41, no.~2, pp. 515--522, 2018.

\bibitem{ines2021biomedical}
A.~In{\'e}s, C.~Dom{\'\i}nguez, J.~Heras, E.~Mata, and V.~Pascual, ``Biomedical
  image classification made easier thanks to transfer and semi-supervised
  learning,'' \emph{Computer methods and programs in biomedicine}, vol. 198, p.
  105782, 2021.

\bibitem{ge2020deep}
C.~Ge, I.~Y.-H. Gu, A.~S. Jakola, and J.~Yang, ``Deep semi-supervised learning
  for brain tumor classification,'' \emph{BMC Medical Imaging}, vol.~20, no.~1,
  pp. 1--11, 2020.

\bibitem{53_label_propagation}
U.~N. Raghavan, R.~Albert, and S.~Kumara, ``Near linear time algorithm to
  detect community structures in large-scale networks,'' \emph{Physical review
  E}, vol.~76, no.~3, p. 036106, 2007.

\bibitem{yamada2022guiding}
T.~Yamada, M.~Massot-Campos, A.~Pr{\"u}gel-Bennett, O.~Pizarro, S.~B. Williams,
  and B.~Thornton, ``Guiding labelling effort for efficient learning with
  georeferenced images,'' \emph{IEEE Transactions on Pattern Analysis and
  Machine Intelligence}, vol.~45, no.~1, pp. 593--607, 2022.

\bibitem{zhou2018brief}
Z.-H. Zhou, ``A brief introduction to weakly supervised learning,''
  \emph{National science review}, vol.~5, no.~1, pp. 44--53, 2018.

\bibitem{jain2009active}
P.~Jain and A.~Kapoor, ``Active learning for large multi-class problems,'' in
  \emph{2009 IEEE Conference on Computer Vision and Pattern Recognition}.\hskip
  1em plus 0.5em minus 0.4em\relax IEEE, 2009, pp. 762--769.

\bibitem{chapelle2009semi}
O.~Chapelle, B.~Scholkopf, and A.~Zien, ``Semi-supervised learning (chapelle,
  o. et al., eds.; 2006)[book reviews],'' \emph{IEEE Transactions on Neural
  Networks}, vol.~20, no.~3, pp. 542--542, 2009.

\bibitem{vovk_algorithmic_2005}
\BIBentryALTinterwordspacing
V.~Vovk, A.~Gammerman, and G.~Shafer, \emph{\BIBforeignlanguage{en}{Algorithmic
  {Learning} in a {Random} {World}}}.\hskip 1em plus 0.5em minus 0.4em\relax
  New York: Springer-Verlag, 2005. [Online]. Available:
  \url{http://link.springer.com/10.1007/b106715}
\BIBentrySTDinterwordspacing

\bibitem{angelopoulos2021gentle}
A.~N. Angelopoulos and S.~Bates, ``A gentle introduction to conformal
  prediction and distribution-free uncertainty quantification,'' \emph{arXiv
  preprint arXiv:2107.07511}, 2021.

\bibitem{zhan2020electronic}
X.~Zhan, Z.~Wang, M.~Yang, Z.~Luo, Y.~Wang, and G.~Li, ``An electronic
  nose-based assistive diagnostic prototype for lung cancer detection with
  conformal prediction,'' \emph{Measurement}, vol. 158, p. 107588, 2020.

\bibitem{liu2021boost}
L.~Liu, X.~Zhan, R.~Wu, X.~Guan, Z.~Wang, W.~Zhang, M.~Pilanci, Y.~Wang,
  Z.~Luo, and G.~Li, ``Boost ai power: Data augmentation strategies with
  unlabeled data and conformal prediction, a case in alternative herbal
  medicine discrimination with electronic nose,'' \emph{IEEE Sensors Journal},
  vol.~21, no.~20, pp. 22\,995--23\,005, 2021.

\bibitem{liu2022cpsc}
L.~Liu, X.~Zhan, X.~Yang, X.~Guan, R.~Wu, Z.~Wang, Z.~Luo, Y.~Wang, and G.~Li,
  ``Cpsc: Conformal prediction with shrunken centroids for efficient prediction
  reliability quantification and data augmentation, a case in alternative
  herbal medicine classification with electronic nose,'' \emph{IEEE
  Transactions on Instrumentation and Measurement}, vol.~71, pp. 1--11, 2022.

\bibitem{zhan2018online}
X.~Zhan, X.~Guan, R.~Wu, Z.~Wang, Y.~Wang, Z.~Luo, and G.~Li, ``Online
  conformal prediction for classifying different types of herbal medicines with
  electronic nose,'' 2018.

\bibitem{wang2022unsupervised}
H.~Wang, X.~Zhan, L.~Liu, A.~Ullah, H.~Li, H.~Gao, Y.~Wang, R.~Hu, and G.~Li,
  ``Unsupervised cross-user adaptation in taste sensation recognition based on
  surface electromyography,'' \emph{IEEE Transactions on Instrumentation and
  Measurement}, vol.~71, pp. 1--11, 2022.

\bibitem{cancer2012comprehensive}
B.~. W. H. . H. M. S. C. L. . . P. P. J. . K.~R. 13, G.~data analysis: Baylor
  College~of Medicine Creighton Chad J. 22 23 Donehower Lawrence A. 22 23
  24~25, I.~for Systems Biology Reynolds Sheila 31 Kreisberg Richard B. 31
  Bernard Brady 31 Bressler Ryan 31 Erkkila Timo 32 Lin Jake 31 Thorsson
  Vesteinn 31 Zhang Wei 33 Shmulevich Ilya~31 \emph{et~al.}, ``Comprehensive
  molecular portraits of human breast tumours,'' \emph{Nature}, vol. 490, no.
  7418, pp. 61--70, 2012.

\bibitem{rehurek_software_2010}
R.~Řehůřek and P.~Sojka, ``\BIBforeignlanguage{English}{Software {Framework}
  for {Topic} {Modelling} with {Large} {Corpora}},'' in
  \emph{\BIBforeignlanguage{English}{Proceedings of the {LREC} 2010 {Workshop}
  on {New} {Challenges} for {NLP} {Frameworks}}}.\hskip 1em plus 0.5em minus
  0.4em\relax Valletta, Malta: ELRA, May 2010, pp. 45--50.

\bibitem{pagliardini_unsupervised_2018}
\BIBentryALTinterwordspacing
M.~Pagliardini, P.~Gupta, and M.~Jaggi, ``Unsupervised {Learning} of {Sentence}
  {Embeddings} using {Compositional} n-{Gram} {Features},'' \emph{Proceedings
  of the 2018 Conference of the North American Chapter of the Association for
  Computational Linguistics: Human Language Technologies, Volume 1 (Long
  Papers)}, pp. 528--540, 2018. [Online]. Available:
  \url{http://arxiv.org/abs/1703.02507}
\BIBentrySTDinterwordspacing

\bibitem{zhang_biowordvec_2019}
\BIBentryALTinterwordspacing
Y.~Zhang, Q.~Chen, Z.~Yang, H.~Lin, and Z.~Lu,
  ``\BIBforeignlanguage{en}{{BioWordVec}, improving biomedical word embeddings
  with subword information and {MeSH}},''
  \emph{\BIBforeignlanguage{en}{Scientific Data}}, vol.~6, no.~1, p.~52, May
  2019. [Online]. Available:
  \url{https://www.nature.com/articles/s41597-019-0055-0}
\BIBentrySTDinterwordspacing

\bibitem{goldman2020visualizing}
M.~J. Goldman, B.~Craft, M.~Hastie, K.~Repe{\v{c}}ka, F.~McDade, A.~Kamath,
  A.~Banerjee, Y.~Luo, D.~Rogers, A.~N. Brooks \emph{et~al.}, ``Visualizing and
  interpreting cancer genomics data via the xena platform,'' \emph{Nature
  biotechnology}, vol.~38, no.~6, pp. 675--678, 2020.

\bibitem{thennavan2021molecular}
A.~Thennavan, F.~Beca, Y.~Xia, S.~Garcia-Recio, K.~Allison, L.~C. Collins,
  M.~T. Gary, Y.-Y. Chen, S.~J. Schnitt, K.~A. Hoadley \emph{et~al.},
  ``Molecular analysis of tcga breast cancer histologic types,'' \emph{Cell
  genomics}, vol.~1, no.~3, 2021.

\bibitem{xu2022tri}
C.~Xu, Q.~Xu, L.~Liu, M.~Zhou, Z.~Xing, Z.~Zhou, C.~Zhou, X.~Li, R.~Wang, Y.~Wu
  \emph{et~al.}, ``A tri-light warning system for hospitalized covid-19
  patients: Credibility-based risk stratification under data shift,''
  \emph{medRxiv}, pp. 2022--12, 2022.

\bibitem{shrunken_centroid}
R.~Tibshirani, T.~Hastie, B.~Narasimhan, and G.~Chu, ``Diagnosis of multiple
  cancer types by shrunken centroids of gene expression,'' \emph{Proceedings of
  the National Academy of Sciences}, vol.~99, no.~10, pp. 6567--6572, 2002.

\bibitem{softmax_MIT}
I.~Goodfellow, Y.~Bengio, and A.~Courville, ``Softmax units for multinoulli
  output distributions. deep learning,'' 2018.

\bibitem{knowledge_distillation}
G.~Hinton, O.~Vinyals, and J.~Dean, ``Distilling the knowledge in a neural
  network,'' \emph{arXiv preprint arXiv:1503.02531}, 2015.

\bibitem{bengio200611}
Y.~Bengio, O.~Delalleau, and N.~Le~Roux, ``11 label propagation and quadratic
  criterion.''

\bibitem{chen2017combining}
X.~Chen and T.~Wang, ``Combining active learning and semi-supervised learning
  by using selective label spreading,'' in \emph{2017 IEEE international
  conference on data mining workshops (ICDMW)}.\hskip 1em plus 0.5em minus
  0.4em\relax IEEE, 2017, pp. 850--857.

\end{thebibliography}

\end{document}